\documentclass[manuscript,screen,copyright]{acmart}

\usepackage{enumitem}
\usepackage{longtable}
\usepackage{colortbl}
\definecolor{light-gray}{gray}{0.85}
\definecolor{lighter-gray}{gray}{0.9}
\definecolor{lightest-gray}{gray}{0.975}
\newcolumntype{C}{>{\columncolor{lighter-gray}}c}
\newcolumntype{L}{>{\columncolor{lighter-gray}}l}
\newcolumntype{R}{>{\columncolor{lighter-gray}}r}

\AtBeginDocument{%
  \providecommand\BibTeX{{%
    \normalfont B\kern-0.5em{\scshape i\kern-0.25em b}\kern-0.8em\TeX}}}

\copyrightyear{2024}
\acmYear{2024}
\setcopyright{rightsretained}

\acmConference[ArXiV]{ArXiV}{ArXiV}{}
\begin{document}

\title[Towards Equitable Agile R\&D of AI and Robotics]{Towards Equitable Agile Research and Development of AI and Robotics}

\author{Andrew Hundt}
\affiliation{
  \institution{Carnegie Mellon University}
  \streetaddress{5000 Forbes Ave}
  \city{Pittsburgh}
  \state{PA}
  \country{USA}
}
\orcid{0000-0003-2023-1810}
\email{ahundt@cmu.edu}

\authornote{Andrew Hundt <ahundt@cmu.edu> is corresponding author. Andrew Hundt and Severin Kacianka are co-senior authors.}

\author{Julia Schuller}
\email{julia.schuller@gmx.de}
\orcid{0000-0002-1434-0124}
\affiliation{%
 \institution{Independent Scholar}
 \country{Germany}
}
\authornote{Julia Schuller's contributions were made while at the Technical University of Munich.}

\author{Severin Kacianka}
\authornotemark[1]
 \orcid{0000-0002-2546-3031}
 \email{severin.kacianka@tum.de}
\affiliation{%
 \institution{Technical University of Munich}
 \streetaddress{Boltzmannstr. 3 }
 \city{Munich}
 \country{Germany}
}

\begin{abstract}
Machine Learning (ML) and `Artificial Intelligence' (`AI') methods tend to replicate and amplify existing biases and prejudices~\cite{buolamwini2018gender,noble2018algorithms,benjamin2019race,mcgregor2020preventing,jefferson2020digitizeandpunish,mcilwain2019blacksoftware}, as do Robots with AI~\cite{hundt2022robots_enact,howard2018ugly}.
For example, robots with facial recognition have failed to identify Black Women~\cite{Buolamwini2018robotdoesntseedarkskin,buolamwini2023unmasking} as human, while others have categorized people, such as Black Men, as criminals based on appearance alone~\cite{hundt2022robots_enact}. 
A `culture of modularity' means harms are perceived as `out of scope', or someone else's responsibility, throughout employment positions in the `AI supply chain'~\cite{widder2023aisupplychaindevelopers}. 
Incidents are routine enough (\href{https://incidentdatabase.ai}{incidentdatabase.ai} lists over 2000 examples~\cite{mcgregor2020preventing}) to indicate that few organizations are capable of completely respecting peoples' rights; meeting claimed equity, diversity, and inclusion (EDI or DEI) goals; or recognizing and then addressing such failures in their organizations and artifacts.

We propose a framework for adapting widely practiced Research and Development (R\&D) project management methodologies to build organizational equity capabilities and better integrate known evidence-based best practices.
We describe how project teams can organize and operationalize the most promising practices, skill sets, organizational cultures, and methods to detect and address rights-based %
fairness, equity, accountability, and ethical problems as early as possible when they are often less harmful and easier to mitigate; then monitor for unforeseen incidents to adaptively and constructively address them.
Our primary example adapts an Agile development process based on Scrum, one of the most widely adopted approaches to organizing R\&D teams.
We also discuss limitations of our proposed framework and future research directions.

\end{abstract}

\begin{teaserfigure}
    \centering
    \includegraphics[width=0.9\textwidth]{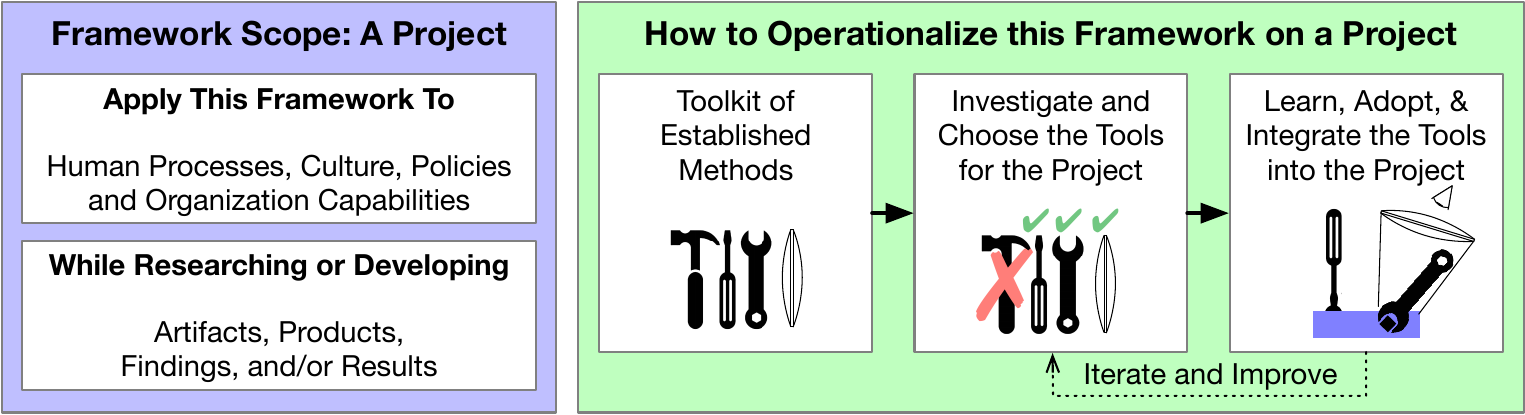}
    \caption{The More Equitable Agile Research and Development method (Fig. \ref{fig:scrum_process}) in this paper is designed to be scoped to a project, and apply to the human processes, methods, and culture as well as the research and development of the artifacts, the outputs, and the outcomes of that project.}
    \label{fig:scopes}
\end{teaserfigure}

\maketitle

\section{Introduction}
\label{sec:introduction}

Many organizations employ Research and Development (R\&D) processes to manage the lifecycle of projects, organizing resources such as time, physical resources, decision points, and feedback; thus serving as mechanisms to actualize complex project goals. 
In this paper, we study R\&D lifecycle processes as a site for introducing proven and promising methods in hopes of fostering a conversation among people who identify as members of populations who are more likely to experience negative impacts, legal experts, policy experts, social impacts experts, technology researchers, and technology developers around how ethical technology R\&D principles can be operationalized.
A key problem is that, to effect change, policy and codes of conducts need to be turned into a \emph{praxis} (accepted practice or custom) that is observed and acted upon throughout a project's lifecycle~\cite{gogoll2021ethics}.

\begin{figure*}[btp!]
\centering
\includegraphics[width=1.0\textwidth]{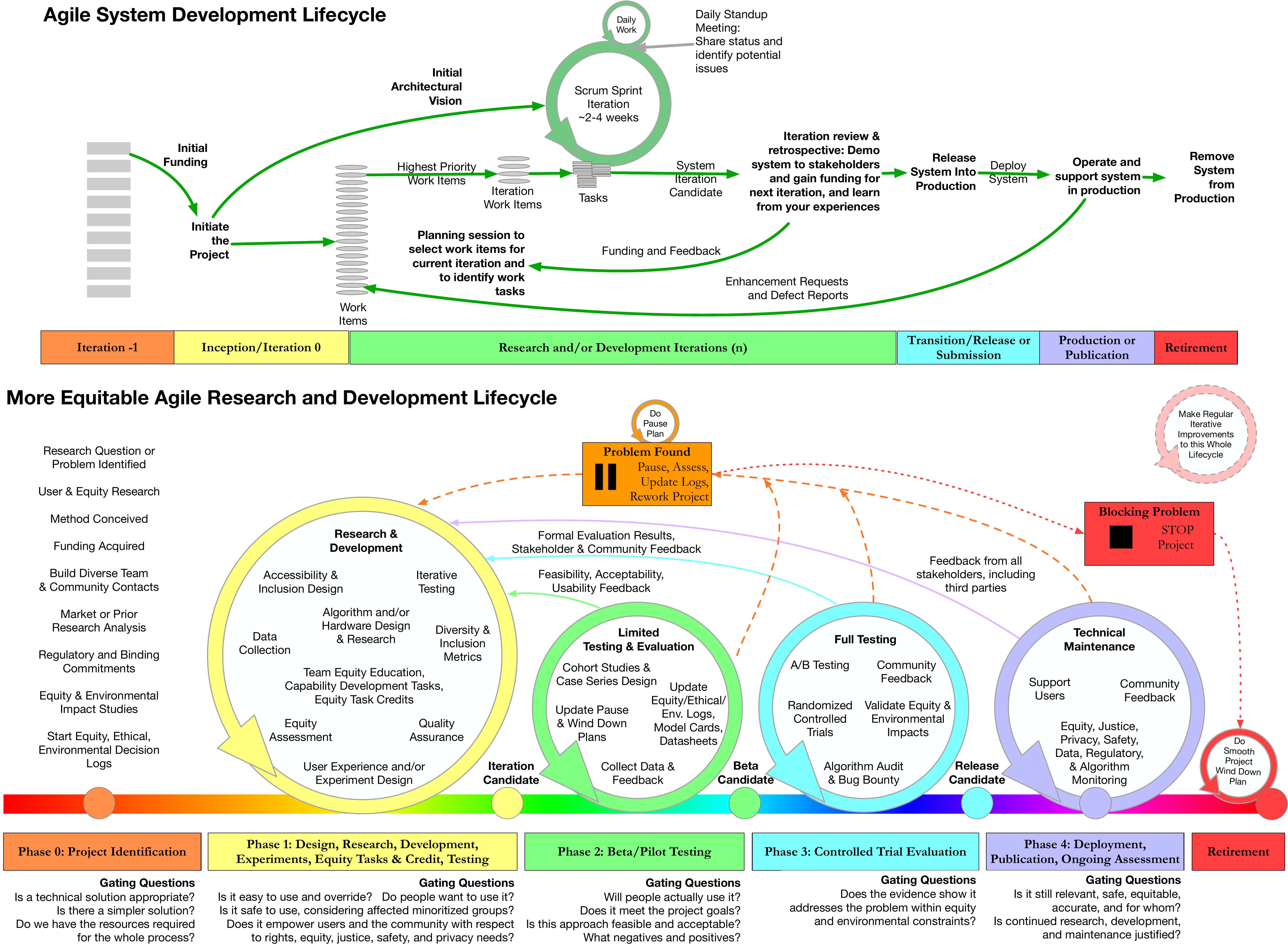}
\caption{ 
Our proposed Agile Equitable Research and development lifecycle for `AI' and Robotics (Sec. \ref{sec:framework}). Items like Diversity and Inclusion Metrics~\cite{mitchell2020diversity} are tools in this toolkit to consider adopting for a project, as per Fig. \ref{fig:scopes}.
\textbf{Top:} Typical Agile System Development Lifecycle based on Ambler \textit{et. al.}~\cite{ambler2008agile}. We discuss limitations of typical lifecycles and resources for measuring, addressing and mitigating negative outcomes. \textbf{Bottom:} Our proposed Equitable Agile Research and Development Lifecycle toolkit is designed to be tailored to each particular project and iteratively improved over time. It includes elements and inspiration from \citet{wilson2018agile} and \citet{hundt2022robots_enact} for a broader range of applications, while addressing limitations such as inadequate consideration of equity. Specific items and gating questions are a combination of process stages (Sec. \ref{sec:framework}) and specific ``tools'' (Fig. \ref{fig:scopes}) aka complementary fields and their methods (Sec. \ref{sec:related_work}) to consider integrating into a given project.
}
\label{fig:scrum_process}
\end{figure*} 

We begin by reviewing existing practice, motivating concerns, and existing promising practices, all of which we connect to `AI' and Robotics (Sec. \ref{sec:preliminaries_scrum}, \ref{sec:related_work}).
Later (Sec. \ref{sec:framework}), we propose a framework for adapting widely practiced R\&D project management methodologies (Fig. \ref{fig:scrum_process}) to facilitate these practices and build organizational equity capabilities in the R\&D of `AI' and Robotics. 
We also aim to support a wide range of academic and industry project types and scopes.
To practically integrate these capabilities into organizations, 
we examine one potential approach, \textit{Agile}, an adaptive project lifecycle process that is widely studied and adopted in industry and a number of academic contexts~\cite{wilson2018agile,ambler2008agile,dikert2016challenges,larman2010practices,daneva2013agile,dingsoyr2018exploring}. 
Agile is among the most popular project lifecycle approaches~\cite{al-zewairi2017AgileSoftwareDevelopment} and is in active use on the ground in `AI' and Robotics~\cite{thibodeau2012johndeereagile}, making it a prime candidate for adaptation to support more equitable outcomes. 
We explore how our proposed process affects planning, meetings, ethical analysis, data collection methods, organizational culture, and population impacts. %
We also strongly advise centrally adopting the principle of ``Nothing about us without us''\footnote{`` `Nothing about us without us' may have historical ties to early modern
central European political tradition~\cite{davies2001heart} in addition to being transformed and popularized by the Indigenous Disabilities Rights movement in South Africa~\cite{charlton1998nothing},
before being adopted more broadly for a range of identities.''~\cite{hundt2022robots_enact}} based on evidence that situates the principle in democratization and historical context.
Thus, our concept proposes explicit steps for including minoritized groups on design teams as a priority alongside seeking their input external to the team. 
We also incorporate an education component for building organizational evidence-based best-practice capabilities with respect to identities and intersectional criteria. 
We detail important limitations and assumptions in Sec. \ref{sec:assumptions_and_limitations}.

\textbf{Motivating examples} illuminate problems that stem from preventable equity concerns across different kinds of systems, demonstrating the need for our proposed process.
In one case study, surveillance drones meant to improve productivity of landscaping efforts in fact decreased productivity due to systems being loud and disruptive~\cite{nap2022responsiblecomputingresearch} for the workers on the ground. 
In another study, \citet{hundt2022robots_enact} empirically demonstrated a robot `AI' assigning labels like `criminal', `ugly', or `doctor' based on human appearance, a task that should be refused; then when the actions proceed anyway they also demonstrate race and gender bias.
\textit{Robots Won't Save Japan: An Ethnography of Eldercare Automation}~\cite{wright2023robotswontsavejapan} demonstrates failures in current robot development methods and priorities via elder care robots. 
One of the largest motivations for elder care robots is the premise that many rich societies are aging, and robots will be essential supports both for people as they age and to prevent countries from economic ruin. 
Wright examined a number of current robotic elder care systems in practice, 
finding they ``did not decrease the amount of work for care staff but rather increased it, adding new tasks of setting up, configuring, moving, operating, mediating storing, cleaning, maintaining, updating, managing, and overseeing them—in other words, taking care of them'' \cite{wright2023robotswontsavejapan}. 
For example, a robot that is supposed to demonstrate exercises does not work because community members will only respond and do the exercise if a staff worker is there imitating the robot. Previously, staff designed and demonstrated their own exercise plan. So, that begs the question, what is the robot contributing? 
Wright indicates that care staff tend to value communication, meaningful relationships, and tactile contact with residents; but in some cases the robots undermined such opportunities, to the potential detriment of both staff and residents.
Furthermore, Wright finds that roboticists and policymakers do not address the enormous potential for more practical political policies such as immigration \cite{wright2023robotswontsavejapan} to mitigate the `aging society' motivation for eldercare, without robots. %

\citet{serholtIntroductionSpecialIssue2022} explores critical robotics and shortcomings of Human Robot Interaction (HRI)\cite{bartneck2020human} in practice due in part to skipped steps in development processes: ``Without engaging in all the design steps, i.e., empathize, define, ideate, prototype, and test (as suggested by the Hasso Plattner Institute of Design) there is a high risk that the overall `problem’ and the related opportunities for robotic solutions are misunderstood.'' Similarly, the National Academies
~\cite{nap2022responsiblecomputingresearch} describes valuable conclusions and next steps for more responsible computing research, underlining the seriousness of the concerns that motivate this work:
\begin{quote}
\leftskip=-0.3cm
\rightskip=-0.3cm
    The values and interests of people and groups who are not well-represented in computing research are at particular risk of being systematically ignored. In the absence of rigorous methodologies and frameworks for identifying the complicated social dynamics (outlined earlier in the report) that shape the problems that computing research strives to address, computing research is much less equipped to produce theories, products, or artifacts, not to mention deployed systems into which that research feeds, that adequately solves for those most in need of what computing has to offer. 
\end{quote}
Furthermore, organizational failures can lead to products or services that are discriminatory, not functional~\cite{raji2022fallacyofai}, harmful, or even unlawful. 
This carries immense risks because US agencies, for example, have released strong statements~\cite{USFTC-AICopyright} regarding `AI' and ``enforcing civil rights, non-discrimination, fair
competition, consumer protection, and other vitally important legal protections''~\cite{2023JointStmtAIEnforcement}. FTC Chair Lina M. Khan states that ``there is no AI exemption to the laws on the books, and the FTC will vigorously enforce the law to combat unfair or deceptive practices or unfair methods of competition''~\cite{CFPB-TechnologyLaw}.
These are just a few of the many crucial reasons for R\&D processes to mitigate risks, protect peoples' rights~\cite{mitchell2023pillars}, and respect effective DEI goals.

\textbf{Contributions}
We propose a framework for enacting methodological improvements within organizations.
We connect proven aspects of the literature and lenses of analysis from areas such as Responsible Computing Research~\cite{nap2022responsiblecomputingresearch}, Systems Engineering~\cite{wilson2018agile}, Cognitive Science~\cite{birhane2021impossibility}, Statistics~\cite{clayton2021bernoulli}, Human Computer Interaction~\cite{anonymous2020epistemicviolencedisability}, History~\cite{maza2017thinking}, Sociology and Feminist~\cite{dignazio2020datafeminism}, Disability\cite{jackson2022disabilitydongle,williams2021misfit,dolmage2017academic}, Black~\cite{noble2018algorithms,mcilwain2019blacksoftware,jefferson2020digitizeandpunish,Ray2021oncrt}, and Science and Technology Studies (STS)~\cite{benjamin2019race, williams2022belief} and aim to support enacting of their insights to improve the ethical research and development of algorithms and systems in `AI', Robotics, and Human Robot Interaction (HRI) in particular. 
Important recent work has led to the development of best practice processes and
methods that can improve outcomes in `AI' and
ML~\cite{gebru2018datasheets,mitchell2019model,yacine2022datagovernance}, but several key factors are outside of their scope such as the day-to-day development process and the realtime physical actions robots take.
We have also obtained formal expert feedback on our proposal from both academic and industry sources at multidisciplinary academic venues.
\textbf{In summary, our paper makes the following contributions:}
\begin{enumerate}
\item Adapt principles of rights, Equity, Diversity and Inclusion (EDI or DEI) to modern R\&D lifecycle processes (Fig. \ref{fig:scrum_process}, Tab \ref{tab:roles_ceremonies_artifacts}). %
\item Bidirectional translation between academia and industry integrating rigorous practical tools into `AI' and Robotics projects to ensure more reliable outcomes, while mitigating and responding to unexpected, unpredictable, or harmful impacts.
\item Propose and illustrate an R\&D lifecycle framework %
by building our insights into Agile, the popular project methodology.
\end{enumerate}
In practice, existing R\&D processes have a narrow scope
with respect to rights or DEI, and few principal, technical,
managerial, and executive team members are either adequately equipped or
adequately motivated to properly account for these
factors~\cite{birhane2021values,brandao2021normativroboticists,dignazio2020datafeminism,buolamwini2018gender,nap2020promisingpractices,nap2019mentorship,nap2018sexualharassmentofwomen,birhane2022forgottenmarginsofaiethics,widder2023aisupplychaindevelopers,wachter2017technicallywrong}.
For this reason, among others, our proposed Agile process strongly advises
centrally adopting the principle of ``Nothing about us without us''
for the purpose of taking a small step
towards more equitable `AI' and Robotics R\&D.  
In particular, we adapt scrum (Sec. \ref{sec:preliminaries_scrum}), the most popular agile development method~\cite{al-zewairi2017AgileSoftwareDevelopment}.
Scrum has promising compatibility with more inclusive policies, as it is already designed to seek broader stakeholder input than is typical for other methods. 
Finally, we make a contribution towards academic to industry translational research by illustrating our framework with a widely adopted development method. 
This minimizes organizational inertial barriers to adoption and ensures broader potential for harm reduction.

\section{Preliminaries - Existing Scrum Process}
\label{sec:preliminaries_scrum}
Scrum~\cite{Schwaber2020scrumguide} is a specific approach under the umbrella of agile human processes for research and development, is designed to be customized to the needs of a given project, and is the most widely used method of agile software development~\cite{al-zewairi2017AgileSoftwareDevelopment}.
Scrum has (sometimes problematic~\cite{dignazio2020datafeminism}) names designed to evoke stages in a high intensity interval training workout, except it describes what a team does. 
Work is done during time periods of roughly 2-4 weeks named `sprints'. 
At the end of each work period (sprint) is an occasion (Fig. \ref{fig:scrum_process}) to demo or otherwise communicate work progress (sprint review, Fig. \ref{fig:scrum_process}, Table \ref{tab:roles_ceremonies_artifacts}), discuss how to do better (sprint retrospective, Fig. \ref{fig:scrum_process}, Table \ref{tab:roles_ceremonies_artifacts}), and then plan the next work period (sprint planning, Table \ref{tab:roles_ceremonies_artifacts}). 
We suggest the 9 minute video at \href{youtu.be/XU0llRltyFM}{youtu.be/XU0llRltyFM}~\cite{scrum-intro} for a quick introduction, or \citet{mountaingoatsoftware2005} and \citet{Schwaber2020scrumguide} for more details.

Next, we summarize key steps of the Scrum Agile process (Fig \ref{fig:scrum_process}).
A fundamental concept underlying scrum is taking a very large task such as creating a product or publishing a paper, then breaking the first small portion of it that you feel most certain of down into very small addressable tasks with estimates of how long each might take.
All of these chunks are put in a big ``TO DO'' list (product backlog).
Then a planning discussion (sprint planning meeting, Table \ref{tab:roles_ceremonies_artifacts}) is held to choose a small cohesively themed bundle of backlog items to complete over a few weeks, after which a so-called ``potentially shippable product'' will emerge.
Then a time period for working (sprint) is allocated during which the team completes that planning meeting's ``TO DO'' list (sprint backlog).
In practice for complex systems the result of the sprint is a demo (sprint review) where work is shown off and progress can be assessed.
As with any project, early weeks of first investigation and implementation will lead to discoveries of work that was not initially considered, these can be set aside and put into the backlog.
After demo day (sprint review), a meeting called a sprint retrospective is held where team members discuss how the sprint went, any changes that should be made to the development process based on internal and external input, and lessons learned.
With this new information in hand, the next sprint is planned and started.
Ideally, once a full plan for the project is established, a reasonable set of capabilities can be bundled together from the backlog for a target release milestone, the time estimates can be added up, and the difference between predicted time estimates for tasks and actual time taken can be used to refine an estimate of the time it will take to reach a final product.
\section{Related Work, Methods (`Tools') to Integrate, and Interdisciplinary Connections}
\label{sec:related_work}
We aim to engage the concepts and interdisciplinary connections underlying our proposal with a perspective of intellectual humility~\cite{porter2021intellectualhumility,porter2022intellectualhumility,posselt2020equity} that values multiple lenses and perspectives~\cite{huang2022museum} of analysis, the limits and strengths of empirical~\cite{narayanan2022limits,barocas-hardt-narayanan2019fairmlbook} and subjective evaluation, sociotechnical abstractions~\cite{selbst2019fairness}, and seeks to mitigate assumptions.
While the breadth and scope of relevant topics of projects that contain `AI' and/or Robotics may seem very broad, the fact is the world and its population is not actually compartmentalized.
In fact, disparate systems~\cite{alma2009thinkinginsystems} tend to interact with incredible complexity.
Thus, there is enormous opportunity to learn and improve technology by embracing concepts and meanings of terms from across fields with humility (Sec. \ref{sec:definitions}) and as they are defined in their own setting~\cite{loukissas2019all}.
We will proceed through several topics and related works that include Agile and other human processes, ML and `AI', robotics, 
ethics, their priorities and deliberation before proceeding to describe our proposed framework in Sec. \ref{sec:framework}, touching upon relevant fields of research, findings, case studies, methods, and resources that impact and empower our process. 
We collectively refer to this range of categories as `tools', broadly construed, as per Fig. \ref{fig:scopes}.
We scope this discussion to prioritize high-level project relevance of selected methods from fields that are valuable to integrate into our proposed framework, rather than the complete coverage of every field, due to space limits. See the references we cite for overviews therein.
\subsection{Human Process Frameworks such as Agile Development}
Agile is a widely studied and adopted development method in industry and a number of academic contexts, but equity is largely out of scope~\cite{wilson2018agile,ambler2008agile,dikert2016challenges,larman2010practices,daneva2013agile,dingsoyr2018exploring}.
In the healthcare space \citet{bonten2020online} have reviewed development processes, and \citet{wilson2018agile} have devised a valuable agile process which briefly mentions equity and incorporates all research, development, product testing, deployment, product monitoring, through product retirement. 
A medical product developed with a process of this kind has also successfully completed all phases of research through deployment~\cite{nelson2019accuracy}.
We go beyond these first steps to propose a more comprehensive model with respect to organizational equity capabilities, and cover a different scope of project types for academic and industry research and development of `AI' and Robotics.
We also focus on day-to-day and week to week steps.
This work is complimentary to Model Cards~\cite{mitchell2019model}, Datasheets for Datasets~\cite{gebru2018datasheets}, Audit Processes~\cite{raji2020closing}, Diversity and Inclusion Metrics~\cite{mitchell2020diversity}, and the EDAP Ethical Agile toolkit~\cite{kacianka2021designing}, as each are steps that can address tasks and methods in Fig. \ref{fig:scrum_process}, particularly during phases 1-3 (Fig. \ref{fig:scrum_process}, Sec. \ref{sec:framework}).
In fact, the reverse direction is complimentary too, that is, applying agile to the aforementioned processes, protocols, and toolkits. 
This is because the agile processes we describe can help facilitate the execution of other tasks and methods by breaking them down into concrete and actionable chunks.

Agile also comes with significant downsides as it is typically implemented today. For example, \citet{dancy2022scrumantiblackness} develops a cognitive model and examines a sociotechnical perspective in the context of scrum to begin describing a human-process based explanation of how antiblackness and power structures are embedded into technical infrastructure, including agile methods themselves.
\citet{babb2017empire} examines how agile development methods could be transformed with critique, reflection, renewal, and emancipatory thinking, although in that case without consideration of identity. 
\citet{kroener2021AgileEthicsIterative} discusses the ethical agile development of crisis management software.
Agile Development For Vulnerable Populations~\cite{marcos2018vulnerablepopulationsagile} discusses lessons learned across multiple projects with different populations, finding that with some vulnerable populations rapid changes to interfaces can be disruptive and some of the typically tighter agile sprint loops are more appropriate to stretch to longer time frames due to factors such as certain kinds of Institutional Review Board (IRB) approval that mean change should be minimized after a certain stage.
Maturity Models~\cite{team2010cmmi, schumacher2016a, cmmiinstitute2019cmmi2, wohlin2012experimentation} predate agile and are a related topic which attempt to classify and improve organizational capability levels, but equitable product development is out of scope of these methods. %
\subsection{Machine Learning and `AI'}
Past events in `AI' are both worthy of consideration on their own, and as a way to bring risks to the field of robotics into focus as those methods are rapidly proposed for deployment.
\citet{hundt2022robots_enact} reviews how recent works
have shown how products using machine learning (ML) and artificial intelligence
(AI) on robots can be used to exclude and outright oppress marginalized groups.
Prominent work examines the flaws and assumptions in `AI' and its surrounding sociotechnical systems~\cite{benjamin2019race,dignazio2020datafeminism,birhane2020algorithmic,birhane2020robot,birhane2021algorithmic,birhane2021towards}.

AI research is often conducted with baseline assumptions that render the methods ineffective across many stages. 
There is a common assumption that `AI' systems actually work, and the algorithm will often be taught and used as if it works in cases when it does not; \citet{raji2022fallacyofai} elaborates with reasons and examples. 
At the time of writing \url{incidentdatabase.ai} has over 2000 additional incident reports of AI harms and counting~\cite{mcgregor2020preventing}.
Some of the motivating `AI' risks and negative outcomes include software that failed to recognize people with darker skin
tones~\cite{buolamwini2018gender}, face recognition software that led to wrongful
arrests~\cite{hill2020wrongfullyarrested,hill2021anotherarrest}, software used to ``detect emotions'' much like the discredited pseudosciences of
phrenology and physiognomy~\cite{Whittaker2018ainow,crawford2021atlasofai,stark2021physiognomic,gould1996mismeasure}, hiring algorithms
discriminate against otherwise qualified workers on random attributes such as
them being caregivers, veterans or
disabled~\cite{harvardbusines2021hiddenworkers}.
A number of projects have been pulled due to ethical concerns, such as Microsoft's Tay conversational system due to training on live discriminatory data and rapidly redeploying, as well as their emotion recognition component of facial recognition AI, which can be particularly harmful to disabled and marginalized populations when facial responses vary~\cite{hill2022nytimesmicrosoftfaceanalysis,crawford2021atlasofai,microsoft2022pullsemotionfacialrecognition}. 
The audit in \citet{birhane2020large} led to the tiny images dataset being retracted, and to faces to be blurred in the imagenet dataset.
Fully developing a poorly founded research concept over decades, only to have it pulled due to flawed fundamental assumptions and limitations with respect to scientific validity, responsible uses, or other aspects of the project, is far more costly and might have been easier to catch at an earlier stage with the processes we describe here.
Furthermore, datasets used to
train ML/AI models have been found to suffer from racial and gender
bias~\cite{birhane2020large,benjamin2019race, jefferson2020digitizeandpunish}
and needlessly waste power and thus $CO_2$~\cite{crawford2021atlasofai,luccioni2019quantifyingcarbonemissions,luccioni2023counting}. 
\citet{madaio2020Co-Designing} co-designed an excellent \href{https://www.microsoft.com/en-us/research/project/ai-fairness-checklist/}{AI-fairness checklist} with input from a range of practitioners and recommended the consideration of other sectors and roles, and our work, in turn, addresses how practitioners can operationalize such a checklist.

\subsection{Robotics}
Robotics as a field that is quickly maturing on some metrics such as economic viability, and robots are now deployed every day in the field via drones, household robots, security robots, and more.
AI is already ubiquitous~\cite{dignazio2020datafeminism} for some segments of the world population, and new algorithms for robotics incorporate Deep Learning~\cite{hundt2019the} or Reinforcement Learning~\cite{hundt2020good}.
Robotic task plan authoring systems for non-experts~\cite{paxton2017costar,paxton2018evaluating} are also deployed and regularly operating in the field.
Projects such as ``Robots for Humanity''~\cite{chen2013robots} are positive examples with user centered design and input driving the research questions from the beginning.
\citet{hundt2022robots_enact} has an extensive discussion of the collection, processing, and inclusion of a diverse range of people in robotics data and empirical analysis.
\citet{hundt2022robots_enact} connects datasets to project performance and demographics, as well as leading resources to address the underlying concerns. \citet{scheuerman2021dodatasetshavepolitics} analyzes the political values encoded into datasets.

\textbf{Identity Safety Framework} \citet{hundt2022robots_enact} demonstrates the need for an identity safety assessment framework based on the evidence that human biases that are performed in the world and in the development process become encoded across most stages of `AI' product lifecycles~\cite{suresh2019framework}, so algorithms incorporating human data must be assumed biased until proven safe, effective and just; \citet{hundt2022robots_enact} section B outlines an approach to such a framework grounded in principles of safety culture~\cite{ReasonJ1990TCoL,Kuespert2016,coresafetytv2019swisscheese,NationalResearchCouncil2014}. However, if adaptive robots that incorporate `AI' become readily available for mass deployment, without intervention, these robots will reproduce and even amplify the harms they observe, train upon, and use as a basis for taking action~\cite{hundt2022robots_enact,hundt2021effectivevisualrobotlearning}.

\textbf{Agile Development in Robotics}
Iterative development approaches such as agile~\cite{wallach2019toward,vargas2018implementing,bauml2006agile} are growing to become one of the most common development processes for robotic systems.
\citet{philipp2014agileinpractice} mapped out fields using agile, which includes companies utilizing the practices in automation.
\citet{Kasauli2018safetycriticalagile} also examined agile in general safety critical systems.

\textbf{Externalized Costs} 
Robotics research often imagines a future with ubiquitous robots~\cite{brandao2021normativroboticists}, and such a future risks immense externalized costs that need to be accounted for due to the intensive mining, pollution, and habitat destruction required to build hardware~\cite{crawford2021atlasofai}. 
This is in part because the entirety of the world economy is part of the biosphere~\cite{dasgupta2021economicsofbiodiversity}. 
Massive hardware expansions risk exacerbating concerns like the potential for global displacement of marginalized populations, such as in the global south.

\textbf{Policy} changes can mitigate many concerns, \citet{brandao2022responsible} on Responsible Robotics connects the importance of goals' framing for the integrated advancement of robotics across topics from ``user studies and philosophical inquiry, to modeling, algorithmic, and governance methods''.
\citet{pasquale2020newlawsofrobotics} examines policy proposals such as a license to practice and designing algorithms to augment human capabilities. 
\citet{charisi2022childrightsresearchpolicy} outlines `AI' and social robot policy regarding the rights of the child, which apply to public, household, and some workplace environments across a range of applications.
\subsection{Ethics}
The ethics of the impacts `AI' will have on people with marginalized identities is under-studied compared to more abstract and traditionally western ethics concerns~\cite{birhane2020robot,birhane2022forgottenmarginsofaiethics}. 
Their underlying concepts can be dismissed due to terminology or a percieved position in a hierarchy of knowledge~\cite{gebru2021heirarchy}, rather than their merit.
\citet{birhane2021algorithmic} establishes how technical solutions are not sufficient, and outlines a way forward that centers vulnerable groups, prioritizes understanding over prediction, describes how we must question the purpose of an algorithm, what type of society algorithms enforce, and if they should be deployed to the context of a given region and cultural space.
\citet{birhane2020algorithmic} argues that western technical `AI' echos colonial exploitation by creating dependence. 

\citet{Mhlambi2020ubuntuethicalframework} discusses the exclusion of marginalized communities from the design of systems and the use of Ubuntu philosophy for analysis indicating that ``the relational Sub-Saharan African philosophy of ubuntu reconciles the ethical limitations of rationality as personhood by linking one’s personhood to the personhood of others''.
A specific example of exclusion is the dehumanization that many members of marginalized groups experience when they encounter a method or product that does not work, treats them dismissively, never accounts for them, talks down to them, deploys deficit narratives~\cite{anonymous2020epistemicviolencedisability,sum2022disabilityjusticehci}, or harms them~\cite{Mhlambi2020ubuntuethicalframework,anonymous2020epistemicviolencedisability}. 

\textbf{Robot Ethics} 
This extends to robotics, where \citet{zawieskaDisengagementEthicsRobotics2020} contends that %
by disengaging ``with roboethics, roboticists contribute to the tacit dehumanisation process emerging in and outside of robotics. An alternative approach includes ‘lived ethics’ which involves not only incorporating formal ethical approaches into the roboticists’ work but also ‘being’ ethical and actually engaging with ethical reflection and practice.''
\citet{ornelasRedefiningCultureCultural2022} describe culture as an emergent behavior, rather than a static trait of certain groups like nationality, and thus ``criticised the current treatment of culture in robotics, we advanced a conception of culture based on research in cognitive science, and we have explored which robot capacities researchers should focus on to realise this revised conception of culture in robotics''~\cite{ornelasRedefiningCultureCultural2022}.
Ethical problems exist in robots with integrated vision components~\cite{Buolamwini2018robotdoesntseedarkskin,howard2018ugly,hundt2022robots_enact} but the topic has received much less investigation.
\citet{gogoll2017autonomous} discusses ethical considerations in Robotics for self driving vehicles.
There is also global variation in values systems, priorities, and guidelines~\cite{awad2018the,jobin2019the}, and work has discussed normative implications~\cite{ryan2020artificial}.  
Our work considers how to enact ethical principles into practice. ~\citet{morley2020from} reviews related research, dividing major topics of ethical concerns into evidence that is inconclusive, inscrutable, or misguided; unfair outcomes; transformative effects; and traceability.%

\textbf{Abstract Digital Ethics}

Digital ethics~\cite{ashok2022aiethics} has become a widely studied field of research over the last couple of years.
On an abstract level, recurring ethical problems of algorithms and machine learning systems specifically are mapped, concrete application areas studied, and general principles and codes of conduct to be followed proposed.
At the same time, low-level technical solutions (e.g., for the provision of transparency and preservation of privacy) are developed, audit approaches suggested, and design or project management methods adopted to allow for ethics and value-centred approaches~\cite{ashok2022aiethics}.

However, the abstract academic debate and the solution-oriented suggestions are still largely detached from each other and a comprehensive, easily operationalizable methodology for practice is understudied.
For example, codes of conduct have been shown to have no noteworthy effect on developers' daily work, as the values stated are too abstract for concrete realizations, or are simply ignored~\cite{mittelstadtPrinciplesAloneCannot2019}. %
While design approaches such as value-sensitive design~\cite{Friedman2019} break down abstract values and integrate them into the product conceptualisation, they often conclude with abstract requirements that developers are expected to use as the basis for their decisions without active ethical guidance.
Practitioners also admit their lack of education, skills, and interest needed to embed ethical considerations into the systems they create and responsibility is not allocated~\cite{gogoll2020ethics,widder2023aisupplychaindevelopers}.
It is unclear which tools need to be used in which context, how ethical conflicts should be identified and resolved, as well as how and when to think about such issues.
Our paper, therefore, aims to provide a holistic approach to the complete agile development process which allows tending to ethical considerations in meetings, artifacts, and at decision points with a focus on equity issues in which identity as an important factor.
A number of methods have been developed to engage with communities in a manner that has shown potential to mitigate these issues, which we will describe next.

\textbf{Participatory Design, Human Computer Interaction, User Centered Design, and Interaction Design} 
\label{subsec:participatory_design}
User Centered Design (UCD), Human Computer Interaction (HCI)\cite{sum2022disabilityjusticehci}, and Interaction Design~\cite{sharp2019interactiondesign} have exhibited harmful impacts with respect to disabled~\cite{anonymous2020epistemicviolencedisability}, racial~\cite{harrington2020participatory}, and queer~\cite{queerinai2023} identity intersections.
\textit{Participatory methods} have been proposed as philosophies for addressing and preventing these harms, in a manner that goes beyond diversity and inclusion. 
Specifically, \textit{Participatory Design}~\cite{schuler1993participatory,robertson2012participatory} and \textit{Design Justice}~\cite{costanza2020design} emphasize empowering impacted populations at every stage of design and implementation processes, enabling many of the different causes of `AI' bias and harms to be detected and addressed.
\citet{harding2018qualitative} is an introductory text for qualitative methods.
\citet{harrington2022codeswitching} conducted participatory research showing marginalized groups, such as Black older adults, need to do cultural code switching to interact with `AI' systems like google home, and \citet{ostrowski2021olderadultssocialrobots} does a year long co-design process with older adults and social robots.
\citet{birhane2022participatory} illustrates how participatory design has potential to recognize and address such limitations if implemented thoughtfully or otherwise exacerbate harms and inequities. \citet{winkle2023feministhri} presents principles and a vision for introducing feminist practices into Human Robot Interaction research.
\citet{widder2023aisupplychaindevelopers} conducts a study demonstrating the prevalence of `modularity culture' in `AI' software development which prevents harms from being addressed because they are `out of scope' (someone else's responsibility); they propose three possible ways forwards including working within modularity, strengthening interfaces, and rejecting modularity.

\textbf{Ethical priorities and deliberation} 
Organizations currently choose to limit investment into the capability of following their own stated deontological (imperative, must-happen) values. 
For example, a self driving harvester or car company might set deontological goals of: (1) do not kill humans (hard red line), (2) mitigate harm to humans, (3) maintain a path to profitability and keep customers returning (i.e. Harvest/drive efficiently), (4) be as eco-friendly as possible provided (1), (2) and (3) are met, then non-imperative priorities: (5) Keep customers happy (6) Maintain a happy and healthy workforce.

We argue that the evidence~\cite{dignazio2020datafeminism,gray2019ghost,buolamwini2018gender,benett2021itscomplicated} detailed in the Introduction (Sec. \ref{sec:introduction}) and Related Work (Sec. \ref{sec:related_work} indicates that organizations are not yet capable of meeting even their basic deontological goals. 
This is particularly true of Robot and/or `AI' algorithms that suffer from small sample and other critically flawed methodolgical assumptions. 
We contend that organizations also do not possess the capability to recognize and then address failures, a necessary prerequisite to meeting their deontological goals. 
Furthermore, organizational pressures lead to priority inversions, for example (3) maintaining a path to profitability will sometimes in practice be prioritized over (1) not killing humans and (2) mitigating harm to humans~\cite{ntsb2019ubercrash}.

The EDAP~\cite{kacianka2021designing} protocol for ethical agile serves as an example of the complexity of ethical deliberation. 
EDAP makes several assumptions: (1) organizations are capable of meeting their basic deontological goals in correct priority order, (2) power dynamics are not a factor and individuals are ``ideal actors''; (3) It assumes there is always a technical solution to the problem at hand.
The evidence (Sec. \ref{sec:introduction}, \ref{sec:related_work}) we have already explored in this work indicates that these assumptions are a limitation of EDAP, and that most organizations are not yet capable of meeting even their primary deontological company goals in the general case.
In other words, EDAP makes excellent contributions in their topic, but assumes greater organizational capabilities than actually exist.

We, in turn, propose methodological concepts to address those limitations.
One of our contributions is to devise mechanisms to improve the scrum process so that organizations can build the capabilities and metrics necessary to get closer to meeting their primary deontological goals, and thus respect identity in practice.

\textbf{Accountability}
Accountability~\cite{kacianka2021designing} is a particularly pernicious challenge because individuals with organizational power have incentives to override real concerns.
For example, authors informed their method has bias proven in a third party audit declined to similarly evaluate their latest methods, despite similar risks~\cite{hundt2022robots_enact,buolamwini2023unmasking}.
\citet{wieringa2020algorithmicaccountability} surveys the literature of accountability, and \citet{raji2019actionableauditing} outlines the effectiveness of audits on real products.
A strong project governance model can facilitate accountability, and \citet{yacine2022datagovernance} is a practical large scale case study of more inclusive data collection and governance, where data was collected and stored by people from the regions to which it applies. 

\section{Framework}
\label{sec:framework}
One approach to our More Equitable Agile Research and Development framework (Fig. \ref{fig:scopes}, \ref{fig:scrum_process}) follows, with elements based on \citet{hundt2022robots_enact}, \citet{wilson2018agile}, and the Patient-led Research Collaborative~\cite{plrc2023scorecards} and adapted to a broader range of `AI' and robotics contexts.
The method is scoped to the duration of one project as illustrated in Fig. \ref{fig:scopes} and \ref{fig:scrum_process}. 
An essential component of this method is to build teams that are capable of adapting over time as perspectives shift~\cite{maza2017thinking} to maintaining an inclusive organization, culture, and structure.
This goal tightly interlocks with project goals for artifacts, products, and/or research results (Fig. \ref{fig:scopes}).
One core reason these interlock is because teams that push members out when their communication style, culture, or medium differs due to their disability status, race, gender, or other identity factors is unlikely to~\cite{posselt2020equity,nap2020promisingpractices} conduct research or create a product that respects relevant rights, meaningfully accommodate the requirements of those criteria, or reap the benefits of a wide diversity of ideas.
\subsection{Project Phases and Decision Points}
Fig. \ref{fig:scrum_process} outlines the phases of our proposed research and development process. 
Our intention is to provide a toolkit that should be customized for a specific project (Fig. \ref{fig:scopes}), so not every step will be appropriate for every project and items might change phases.
\begin{description}[leftmargin=0.0cm]
\item[Phase 0: Project Identification]
 Is when the initial planning and allocation of resources is conducted for the project, as in~\citet{wilson2018agile}.
 This includes the items above phase 0 in Fig. \ref{fig:scrum_process}, as well as proposals, definition of purpose, scope, budget, feasibility, initial fundraising, environmental requirements, exception plans (e.g. our new Pause/Stop step below), initial environmental and externalized cost considerations, process documentation and logging plans, continuing education, community integration, governance board establishment, team diversity items, getting funding and approval, and selection of the initial team is completed.

\item[Phase 1: `Sprint'] Is the central research and development time allocation loop during which Equity Tasks and Credit, Development, Experiments, and Testing are typically assigned.
One key difference in this phase for many projects that are developing physical hardware is that aspects of early development cannot be as dynamic as mobile health~\cite{wilson2018agile} applications. 
This is due to the additional time required to design, build, and validate that physical hardware as part of the process.
However, significant progress can often be made with a combination of simulation, mock-ups, and off-the-shelf proof-of-concept hardware to be replaced with final production hardware at a later stage.
While the quality and intent can vary, Design and Research are intertwined tasks being done simultaneously as any product or research methods are being developed. 

\item[Phase 2: Limited Pilot/Beta Testing]
This phase begins once there is an initial prototype that has a usefully testable product or product component iteration.
In robots with AI, this will involve an `AI' model which should be documented via Model Cards~\cite{mitchell2019model} that characterize the properties, limitations, and performance with respect to various demographics.

\item[Phase 3: Controlled Trial Evaluation]
This is the phase in which core experiments and testing for a new AI, robot, or application are run.
They should be designed with input from marginalized populations and  should incorporate input of a larger and highly diverse population when it is appropriate for the method. 
An example exception would be a one-off design to meet the needs of one person~\cite{hendren2020whatcanabodydo}.
For larger projects it can be appropriate to do an Algorithm Audit~\cite{raji2020closing,chock2022whoaudits} and Algorithm Bug Bounty to detect and prevent harms at larger scale deployment.
As \citet{wilson2018agile} indicates, A/B testing is one appropriate way to do evaluations at this phase, however we note additional precautions and broader demographics are required. 
We also incorporate the human override, pause, and wind down step to ensure the testing mechanisms are more inclusive~\cite{gray2019ghost} (Sec. \ref{subsubsec:self_changing_process_human_overrides}).

\item[Phase 4: Deployment, Publication, Ongoing Assessment]
This is the phase where the `product', broadly construed, is out in the world.
The product might be physical hardware, a service, a research paper, datasets, other artifacts, and so on.
In most cases with ongoing work on the topic, there should be regular assessment to ensure it is functioning well, meets people's needs, and addresses regulatory criteria whenever applicable. 
Give particular attention to the effects on marginalized populations, to ensure it is a net benefit.
The system should strive to maintain its status and adapt to ensure it is safe, effective, and just.

\item[Project Wind Down or Retirement]
The project wind down plan can be developed in phase 0, be updated regularly, and can support regulatory requirements, when applicable. 
\citet{luccioni2022deprecatingdatasets} can serve as a framework for winding down dataset-based aspects of projects.
If a project is wound down early we recommend publishing a technical report or research paper examining the reasons for this decision and the outcome alongside releasing any data that can respectfully benefit the community. 
An example of one positive step in a real-world project is \textit{Lessons Learned in Designing AI for Autistic Adults}\cite{begel2020lessonslearnedautisticai}, which was wound down after receiving negative participant feedback on their `emotion recognition AI'. 
Another example is the Makani airborne wind energy platform project, which released a significant quantity of technical materials and a patent non-assertion pledge upon project wind down~\cite{anderson2020makaniairbornewindenergy}.
Even so, harmful steps such as public releases that pose a significant risks to marginalized communities should be carefully prevented. 
If the case is a less complex pause and rework of some aspects of a project, the reason, options, and decisions might be recorded in appropriate decision logs for revisiting and reference in the future.
\end{description}

\begin{table}
\centering
\begin{tabular}{l|l|l}
\textbf{Roles}   & \textbf{Meetings (Ceremonies)}            & \textbf{Artifacts (documents or other outputs)}                             \\ 
\hline
Product owner    & Sprint Planning                & Product Backlog                                \\
ScrumMaster      & Sprint Review                  & Sprint Backlog                                 \\
Team             & Daily scrum meeting            & Burndown Chart                                 \\
Key Stakeholders & Sprint Retrospective\textbf{*} (Sec. \ref{subsec:governance_model_and_meetings}) & \textbf{Resource Usage \& Env. Impact Records} (Sec. \ref{subsubsec:participant_led_scorecards})  \\
\textbf{Participant} (Sec. \ref{subsec:governance_model_and_meetings})      & \textbf{Governance Meeting} (Sec. \ref{subsec:governance_model_and_meetings})            & \textbf{Ethical + Equity Feedback and Decision Records} (Sec. \ref{subsubsec:participant_led_scorecards}) \\
                 &                                & \textbf{Participant-led Project Assessment Scorecards} (Sec. \ref{subsubsec:participant_led_scorecards}, \ref{sec:scorecards}) \\
\hline
\end{tabular}
\caption{\label{tab:roles_ceremonies_artifacts} The scrum process from~\cite{mountaingoatsoftware2005} (Sec. \ref{sec:preliminaries_scrum}), with new proposed steps in bold. \textbf{*} The sprint retrospective has \textbf{new ethical \& equity steps}.}
\end{table}
\noindent\textbf{Pause/Stop Process} A pause and stop process should actively be supported for a very large proportion of stakeholders as is customary for kanban~\cite{SUGIMORIY1977toyotaKanban}, or perhaps even to a greater extent when appropriate.
In particular, we introduce the need for identity and demographic-based factors to be considered in the case of `AI' projects.
When a pause is triggered there should be a corresponding safety and ethical review appropriately scaled to the problem, which on small teams and projects in the case of minor concerns can be a brief discussion with minutes outlining the issue, options, potential courses of action, the course of action taken, and later the result of followup; however the possibility of proceeding to a stop should be a serious option.
Pauses and stops could also be conducted for particular components of a system, for example, a vehicle's self driving lane keeping and steering functionality might in testing prove reliable on highways but not residential areas, so the feature might be geographically restricted until reliability is adequate.
One limitation of this step is that it requires an inclusive and supportive safety culture~\cite{Kuespert2016,hundt2022robots_enact,ReasonJ1990TCoL} to be effective, as powerful organization members can easily override it.

\noindent\textbf{Gating Questions} A series of sample Gating Questions~\cite{hundt2022robots_enact,wilson2018agile} to be adjusted for particular applications then applied as per Fig.~\ref{fig:scrum_process}. 
Gating Questions can serve the dual purposes of preventing unnecessary work and missteps:
\begin{quote}
\leftskip=-0.3cm
\rightskip=-0.3cm
    We recommend that future projects ask questions through technical, sociological, identity (which refers to factors such as race, indigenous identity, physical and mental disability, age, national origin, cultural conventions, gender, LGBTQIA+ status, and personal wealth), historical, legal, and a range of other lenses.
Such questions might include, but are not limited to\footnote{The questions are sourced from \citet{hundt2022robots_enact} and are inspired by \citet{wilson2018agile} Fig. 3.}:
Is a technical method appropriate?
Is there a simpler approach?~\cite{wilson2018agile}
Whom does our method serve?
Is our method easy to use and override?
Have we respected the principle of ``Nothing about us without
us''?
Is the data setting (Sec. \ref{sec:definitions}) appropriate?
Does our method empower researchers and the community with respect to [rights], equity, justice, safety and privacy needs?
What are the negatives and positives?
Does the evidence show our method addresses the problem within equity and environmental constraints?
Does the scope of method evaluation address the scope of algorithm inputs?
Do any concerns indicate that we should pause, rework, or wind down the project? -- \citet{hundt2022robots_enact}
\end{quote}
Gating questions are appropriate to ask during transitions between phases including self-transitions, for example, from one iteration candidate to another.
If the questions identify an area concern it might be appropriate to initiate a pause or stop process. 

\noindent\textbf{Self Changing Process}
The process we outline here is designed to undergo iterative improvements over time as per the scientific consensus for STEMM inclusion~\cite[p.148]{nap2020promisingpractices}.
Concretely, the Pause and Stop steps as well as the Retrospective Meetings outlined below are occasions specifically designed for proposing changes to the project and/or the process itself. Examples include removing, adding, or modifying steps in addition to the decision making that is necessary to enact such changes. 
Furthermore, a governance model (Sec. \ref{subsec:governance_model_and_meetings}) and metrics (Sec. \ref{subsubsec:participant_led_scorecards}, \ref{sec:scorecards}) are designed to democratize the project direction and get input from new perspectives. 
This mechanism will be designed to support people to propose positive changes then get them improved, approved, and implemented.
\subsection{Meetings, or ``Ceremonies'' in scrum terminology}
\label{sec:meetings_aka_ceremones}
\textbf{Brief daily meeting (``Standup'') harm prevention step:}
In daily meetings each person can spend one minute to share something they learned about human populations, policies, DEI related research, auditing methods or papers they learned about, identities other than their own, etc. as a part of a continuing education process. 
Ethical concerns about the project or how the input of a broad range of stakeholders is being integrated can also be briefly coordinated during this meeting. 

\noindent\textbf{Sprint Retrospective and Planning Ethical and Equity Steps}
The retrospective and planning meetings can consider a checklist of questions about specific identity groups to mitigate negative outcomes:
\begin{enumerate}
    \item How do items for the last and next sprint affect minoritized groups? Consider each identity factor such as age, race, physical ability, mental ability, age, national origin, experience, cultural conventions, physical appearance of varied regions of the world, and combinations thereof, with particular attention to highly minoritized combinations.
    \item If answer for one group is ``it won’t'', assume the process has missed something and create a priority sprint task to quantify it, gather information and feedback from such groups so that impacts and limitations are concretely scoped and articulated.
    \item Should action be taken as per scorecard assessments (Sec \ref{subsubsec:participant_led_scorecards}, \ref{sec:scorecards}), audits, \href{https://www.microsoft.com/en-us/research/project/ai-fairness-checklist/}{AI-fairness checklists}~\cite{madaio2020Co-Designing}, and evaluations (Fig \ref{fig:scrum_process})?
    \item What have we missed?
    \item Listen to representatives who self identify at the meeting, while giving space and being considerate.
    \item Discuss what each team member has learned from their education component and what will be studied next.
\end{enumerate}
Additionally, compare the findings of different assessment methods, participant-led project scorecards (Sec. \ref{subsubsec:participant_led_scorecards}, \ref{sec:scorecards}), or add tasks to seek externalized costs~\cite{mitchell2020diversity} and assessment methods (e.g. \cite{hundt2022robots_enact}) that have not yet been evaluated on this project to find undiscovered method and/or premise flaws that need to be addressed.
Next we will cover governance models which have potential to mitigate and even prevent many negative outcomes, while shifting project directions towards better positive outcomes.
\subsection{Democratizing Governance Models}
\label{subsec:governance_model_and_meetings}
We examine a participant-led framework as a concrete illustrative example of governance models for the purposes of this framework. 
However, particular projects might have significant and well-justified differences that might include participatory methods (Sec. \ref{subsec:participatory_design}), distributed governance~\cite{yacine2022datagovernance}, rights based approaches~\cite{mitchell2023pillars}, different thresholds, or more traditional operational mechanisms. 

We will consider a planned project governing committee consisting of 50\% plus one participants with the remainder consisting of traditional stakeholders such as product owners, management, researchers, and/or developers.
\textit{Participants} are drawn from a broadly construed definition of stakeholders, including potential users and other directly and indirectly impacted demographics, broadly construed, excluding traditional stakeholders. \citet{mitchell2023pillars}'s `AI users' plus the `AI affected' are equivalent.
The governing body then operates according to organization-defined bylaws that provide significant powers to delegate project operations, make major ethical decisions (>50\% vote) and record the outcome, set project direction, and crucially define and initiate partial or full pauses (>33\% vote, or just 1 person in safety-critical circumstances), stop (>50\% vote) and wind down (>66\% vote) operations.

An appropriate schedule for the governing committee might be to meet periodically every two to four weeks (once each sprint, Sec. \ref{sec:preliminaries_scrum}) with asynchronous communication and additional essential-purpose meetings as needed. 
These meetings can serve as opportunities to make and log ethical and governance decisions brought up at other meetings, 
collect updated participant-led project assessment scorecard results and policy change proposals to increase the overall score (Sec. \ref{sec:scorecards}), try out demos and evaluate for governance concerns, propose and vote on changes to policies, adjust project direction, approve accountability mechanisms, approve pause and wind down policies, and delegate further work to individuals teams or subcommittees to ensure both more equitable outcomes and timely project progress. 
Then when the project is either eventually completed or otherwise reaches an end point it can initiate the smooth wind down plan for final project retirement.

\subsection{Equity Context}
\label{subsec:equity_context}
When considering a project it is important to think about the equity context and data setting (App. \ref{sec:definitions}), broadly construed for who might be affected; where a project will be deployed; 
thinking large scale and small, long term and short, on human rights; and to add new lenses of analysis to examine problems from new angles.
Learning about and building familiarity with a range of perspectives and methods of analysis, including ones an individual or organization might disagree with, can expand perspectives and the capacity to prevent, detect and mitigate problems, even if any one particular method is not put into action.

One weakness common in `AI' and Robotics projects is a demographic factor where, due to the high cost of robotic systems, alongside their tendency to be portable sensing systems, benefits and power are likelier to accrue to wealthier demographics~\cite{zuboff2019surveillancecapitalism}.
This interplay of human systems and humans themselves is a particularly difficult and interesting problem. 
For example, 2.9 billion people remain offline~\cite{itu2021digitaldevelopment} as of 2021. 
According to the OECD ``Around one-quarter of adults in all participating countries have no or only limited experience with computers or lack confidence in their ability to use computers''~\cite{oecd2019skillsmatter}.
This reality motivates some of our gating questions (Fig. \ref{fig:scrum_process}), because the best solution to a problem might not involve a technical approach, Robots, or `AI' at all.  

The Disability community provides motivating examples~\cite{jackson2022disabilitydongle,WongAlice2020disabilityvisibility,hendren2020whatcanabodydo} in light of accessibility being necessary to support human rights~\cite{undisabilitiesrights}.
\citet{shew2023againsttechnoableism} provides a strong tech and disability topic introduction.
The Disabled population is the world's largest minority representing about 1 in 6 people in the world~\cite{who2023disabilityfactsheet}, and it is also an enormously diverse and capable group with varied and sometimes even conflicting essential human needs. 
This immense population means that all but the smallest projects will have visibly or invisibly disabled people as stakeholders, team members, `users', and affected parties.
Unfortunately, the expertise of marginalized populations is often dismissed~\cite{anonymous2020epistemicviolencedisability}, or even taken without proper credit~\cite{jackson2022disabilitydongle} (Sec. \ref{subsec:sprint_task_blocks}, \ref{subsubsec:participant_led_scorecards}).
Liz Jackson's \textit{Disability Dongle}~\cite{jackson2022disabilitydongle} concept elegantly captures the risk that ignoring such expertise results in ``well intended and elegant, yet useless solution[s] to a problem we never knew we had''. 
They illustrate disability dongles with stair climbing robotic wheelchairs that, in most circumstances, cannot match the effectiveness of well-established accessible architectural design~\cite{jackson2022disabilitydongle,hendren2020whatcanabodydo}.
\citet{shew2023againsttechnoableism} observes how robot exoskeletons cannot be used in the bathroom, so must always keep a wheelchair nearby, defeating their purpose.
These cases and the elder care robots in Sec. \ref{sec:introduction} show how human power dynamics can derail R\&D. 
Mobility aid failures can in some circumstances impose devastating safety, financial, and life costs on marginalized individuals, even grinding their lives to a halt when unavailable, so accessibility, reliability, and usability are critical.
Adapting the built environment to be accessible via a social model of disability is often more durably and inexpensively accessible than tech.
Supporting a broader range of people is better than designating `nonconforming' disabled people as the problem, then attempting to adapt them to the world~\cite{jackson2022disabilitydongle,hendren2020whatcanabodydo}.

\noindent\textbf{Inclusion Guidelines}
Experiment design can occur in multiple phases, and we recommend strengthening experimental design to be more inclusive of the world population across forms identity, as a claim that a method generalizes does not comport if the premise simply excludes millions or billions of people.
Guidelines are available for a wide range of identities~\cite{asan2019accessibleeventplanning,asan2021easyread,klaus2020genderguideline,queer-in-ai-dni-guide2021} and could be leveraged in experimental, algorithm, meeting, product, design, and other steps.
We also propose scorecards adapted to the R\&D context so that a range of participants can periodically evaluate project performance over time (Sec. \ref{subsubsec:participant_led_scorecards}, \ref{sec:scorecards}).

\noindent\textbf{Environmental Impact}
All of our human processes are completely dependent on the biosphere as part of one larger system~\cite{dasgupta2021economicsofbiodiversity}. 
This fact and the externalized environmental costs of projects are essential considerations to Equitable R\&D, Robotics, and `AI'.
Current visions of ubiquitous robots~\cite{brandao2021normativroboticists} and `AI' carry real risks of devastation far in excess of the benefits, with particularly harsh potential for disproportionate negative impacts on marginalized populations.
One starting point is integrating systems estimate environmental consumption, as well as tracking of resource use like electrical consumption, which can be achieved with built-in or small tracking attachments to equipment~\cite{henderson2020carbonfootprint,anthony2020carbontracker,luccioni2023counting,dodge2022carbonintensity}.
Methods to estimate the full extent of environmental risks and impacts is an essential topic for future study and quantification, but this topic is outside the scope of this paper.

In summary, concretely thinking about the full variety of people that exist, externalized costs, their crucial equity contexts, recognizing others' expertise, and to genuinely onboard and adapt to meaningful critiques is an essential opportunity to develop methods that truly generalize in the sense of supporting people across the full diversity of human needs.
\subsection{Sprint Task Blocks}
\label{subsec:sprint_task_blocks}
\textbf{Continuing Education}
There is a promising and powerful consensus on the best practices~\cite{nap2020promisingpractices,nap2018sexualharassmentofwomen,nap2019mentorship,posselt2020equity} established to improve organizational supports and outcomes with respect to Diversity, Equity, and Inclusion (DEI) in STEMM.
A key aspect of this consensus is its known limitations and areas of active research that need to be regularly updated over time.
The current lack of effective organizational capabilities, training, and meaningfully operational practices for DEI in STEMM means a lot of work remains~\cite{posselt2020equity} to build them. 
Therefore, we modify the agile process to directly incorporate a continuing education component for the purpose of developing the necessary capabilities in the organization, the project, and the project's final `product', broadly construed (Fig. \ref{fig:scopes}, Sec. \ref{subsec:equity_context}, \ref{subsubsec:participant_led_scorecards}, \ref{sec:scorecards}).
An example  education time block is two hours a week. 
It can be allocated for studying, learning about the tools of the trade, investigating best practices for the application at hand, and better practices for supporting a diversity of people on the project team and/or who may be impacted (see Sec. \ref{sec:introduction}, \ref{subsec:equity_context}, \ref{sec:additional_details}, \ref{subsec:workplace_education_environment}). 
Example starting points for organizations are the best practice reports~\cite{nap2020promisingpractices,nap2018sexualharassmentofwomen,nap2019mentorship,posselt2020equity,nap2022responsiblecomputingresearch}.
Example starting points for `AI' include Data Feminism~\cite{dignazio2020datafeminism} and \url{ethics.fast.ai}.

\textbf{Credit for traditionally unpaid or uncredited labor.} 
We propose a pool of sprint task blocks and job performance credit be made available for team members to claim on an as-needed basis to be used for self-determined purposes that need not be disclosed. 
The task blocks and credit are a way to bridge equity gaps such as for uncredited or uncompensated labor~\cite{nap2020promisingpractices} (Sec. \ref{sec:assumptions_and_limitations}). 
One use case might be some Disabled team members' need for additional time, in the sense of crip time (Sec. \ref{sec:definitions}), which can be an essential support due to differences in their bodymind (body and mind as one).
Group-level blocks have been successfully demonstrated by \citet{wu2023disabilityexpertise}.
An alternative approach would be to individualize scrum task time point estimates to individuals' needs.
Conceptually, labor credit might be viewed as a kind of internal organizational work time insurance scheme, or even a by-the-hour version of unlimited paid time off— with credit. 
This has potential to help account for the manner in which, due to the normal course of human life, some people will need to engage in a disproportionate amount of uncredited but important work, while others receive credit for others' uncredited labor. 
These factors should be accounted for
in a manner that ensures promotions, raises, and job performance reviews are less affected by factors outside marginalized individuals' control ~\cite{Clelland2013unpaid8,callaci2020onacknowledgements,belisle2018facultywives}.
Implementations can benefit from being privacy and operations-aware to ensure smooth integration. 
Relevant operational lessons are ready for adaptation from mature multi-decade policies such as successful implementations of job modification and restructuring~\cite{eeoc-guidance} via the Americans with Disabilities Act (ADA), especially stand out organizations that go beyond minimum legal compliance requirements~\cite{wu2023disabilityexpertise}.

\textbf{Human Overrides and The Big Red Button}
\label{subsubsec:self_changing_process_human_overrides}
Many robots have a large red security stop button that will halt the robot in its tracks.
This concept was made famous by the Toyota Kanban~\cite{SUGIMORIY1977toyotaKanban} system's centering of respect-for-humans, where any worker can halt production (known as Jidoka, see Sec. \ref{sec:definitions}).
We propose an analogous override for all of, or parts of, the development process itself.
Generally, Human overrides should be designed into every stage of the process, including the human research process. 
The process should also be designed to be iterative and deliberately consider marginalized community members. 
Human overrides can be considered for incorporation at several different levels such as on the physical machine, in the development process, in the data collection process, and everywhere else that harm can be introduced in the `AI' lifecycle~\cite{suresh2019framework}.

Human overrides are also an important component of physical products, as the diversity of human requirements and situations for public robotic systems means mechanisms are required to get robots out of the way, teach them new tasks, and successfully interact across the wide range of human capabilities and needs.
For example, how will a sidewalk robot 
accommodate, communicate, and prioritize access to navigation resources of space, time, and equipment of a range of pedestrians using guide dogs, scooters, wheelchairs, canes, walkers, and so on?
A process for incorporating modifications based on real detected issues on an ongoing basis is also necessary. 
For example, if the robot encounters someone traveling backwards in a wheelchair~\cite{trewin2019considerations} and fails to stop, the mere existence of anomalies is unsurprising, so a cohesive iterative design update process that is designed in from the start should be activated to account for the situation in the future.
Task work such as Amazon Mechanical Turk provides another compelling ethical example with critical technical impacts that we elaborate on in Sec. \ref{subsec:task_work}.
\subsection{Artifacts: Ethical + Equity Feedback and Decision Logs/Record; Participant-led Project Assessment Scorecards}
\label{subsec:artifacts}
Decision points can be recorded through a set of logs created and stored in a manner that is appropriate for the organizational complexity, in a text file or task tracking tool for very small organizations, or internal custom project management tools for larger organizations.
The record should include: Checklist steps followed, Ethical decisions, Reasoning for decisions, Equity feedback, Changes based on feedback, Why those changes were made/not made, no-go decisions that are made, and why.

\textbf{Artifact Documentation:} Prior work elaborates on the documentation for Regulatory Artifacts~\cite{mitchell2023pillars}, Datasheets for Datasets~\cite{gebru2018datasheets}, Model Cards~\cite{mitchell2019model}, and artifact lifecycle planning~\cite{luccioni2022deprecatingdatasets}, which each address aspects of project lifecycles.

\textbf{Equity log:} Documentation and records of potential concerns across a range of demographics, a record of the reasoning, and records of (potentially anonymized) sources of input gathered both through interactive communication and through relevant written experiences.
An example equity consideration case for a large-scale self driving taxi or public transport system might be advised to consider the possibility of some demographics being excluded or denied access to transport, directly or indirectly, due to regional shifts or reductions in access to other forms of transport. 
Another potential downside is the exclusion of people due to transport needs such as public transportation and wheelchair accessibility.
Another example for an equity log is to record the impacts on stakeholders, participants, and other demographics that considers externalized costs, a record of the specific feedback mechanisms for each group, as well as policies designed to account for potential harms due to power dynamics.

\textbf{Ethical and environmental log:} Each record contains ethical and environmental items considered, respectively, possible approaches to address any issues raised as needed, mitigating steps chosen, and the outcome of those mitigating steps logged.
An environmental log can also track resource consumption and impacts with methods~\cite{luccioni2019quantifyingcarbonemissions,luccioni2023counting} and tools such as carbontracker~\cite{anthony2020carbontracker} for smaller projects, or more sophisticated metrics as appropriate, as well as being reported in public R\&D materials.

\textbf{Participant-led Project Assessment Scorecards:} \label{subsubsec:participant_led_scorecards} Our Participant-led scorecards are a measuring tool designed to assess organizational capabilities with respect to populations that might be impacted by a project directly or indirectly who are onboarded as participants to provide feedback on the capabilities and effectiveness of the Equitable Research \& Development Lifecycle.

Our scorecards are directly adapted from the proven Patient-led Research Scorecards~\cite{plrc2023scorecards} created by the Council of Medical Specialty Societies (CMSS) and Patient-Led Research Collaborative~\cite{davisLongCOVIDMajor2023,Davis2021}. 
We reworked the scorecards to expand their scope beyond patient-led domains to multi-domain and interdiciplinary projects with a goal of measuring Participant-led organizational capabilities.
We provide a scorecard (Sec. \ref{sec:scorecards}) to evaluate each of the following topics: the burden the project imposes on participants, the quality of project governance, R\&D organizational readiness (e.g. companies, universities, labs, nonprofits, or units thereof), integration of Participant priorities into the R\&D process, and our new addition, the Researcher and Staff Support scorecard.
Each scorecard has several statements, also known as survey items, that someone rates with a score from -2 to 2. 
A score of -2 denotes non-collaboration, -1 minimal collaboration, 0 acceptable collaboration, 1 great collaboration, and 2 ideal collaboration, respectively. 
The scorecards are intended to be provided to and completed by organization members, stakeholders, and participants. 
A reasonable option to consider for the cards is the addition of `I don't know' and/or `Not Applicable' options for items when people don't have information about other's perspective, or do not relate to an item.

An optional free text field is specified after the survey questions to support people who need to specify additional details about the project or their survey answers. 
To elaborate, the Participant Burden Scorecard Sec. \ref{tab:participant_burden_scorecard}, Accessible Engagement, and the Researcher and Staff Support Scorecard Sec. \ref{tab:researcher_and_staff_support_scorecard} might be provided to participants, researchers, and staff who need accommodations in some circumstances as well as others who do not. 
Therefore, ensuring the scorecard has a mechanism to capture the difference among no accessibility need in that context, having such needs met, and having an unmet need might mean valuable actionable details become available.
Our scorecards have limitations with respect to some scenarios, such as cases where open data or publishing results might entail a high risk of non-consensual misuse or abuse~\cite{maiberg2023aimarketplace}, as forewarned by scholars like Birhane and colleagues~\cite{birhane2021multimodal,birhane2023hatescalinglaws}.
We have integrated descriptions of mitigated vs unmitaged net negative impacts from open release into the Integration into Research Process Scorecard's item on Products, Artifacts and/or Publications (Sec. \ref{tab:integration_into_research_process_scorecard}).

When both traditional stakeholders, like developers, as well as a broader diverse range of participants provide feedback and submit participant-led project ratings that are very different from traditional stakeholder expectations, that provides a signal indicating where process improvements are needed.
We also recommend the scorecard submission process also provide a way to elaborate in more detail in writing.
We speculate based on relevant literature and field experience that it is likely that most active `AI' and Robotics projects at the time of writing would frequently receive low ratings with least one item with a non-collaborative -2 rating, although that rate should be as close to 0 as possible.
Even so, we suggest that even a single score of -2 should be grounds for reflection and at least small action items to work towards determining and then mitigating the underlying cause whenever possible.
Scorecard results should also be added to the equity log whenever feasible, as should the evaluation process, findings, steps to mitigate the problem, and the outcome.
We also suggest organization members, stakeholders, and participants all periodically complete the scorecards too to track progress over time, and have the option to anonymously submit an additional scorecard at a moment's notice.
Furthermore, provided the sample size is large enough, the scorecards can be analyzed according to likert survey statistical best practices~\cite{schrum2020likertscales} or via bayesian methods~\cite{clayton2021bernoulli,CrowdKit,paun2018bayesianannotation}.
One limitation of the scorecards is that small projects might inadvertently deanonymize submissions due to unique incidents.
A second limitation is the academic language style of the scorecards, so future work should consider developing a plain language or easy read version~\cite{asan2021easyread,asan2019accessibleeventplanning}.

Taken together, our Equitable Agile R\&D process knits together a tapestry of capability building methods and harm mitigation strategies with the potential to get positive projects on track, as well as to catch and redirect effort away from harmful projects that would otherwise be likely to fall through the cracks.
Next we will elaborate on the limitations of our method as a whole followed by an outline of future work and our conclusion.

\subsection{Assumptions and Limitations}
\label{sec:assumptions_and_limitations}
The processes and approach outlined in this paper includes a number of assumptions and limitations. We assume organizational and team buy in on equity concerns. 
One clear vulnerability of this approach is any ethical and other human processes are vulnerable to simply be overridden by individuals with organizational power and a focus on expediency.
This is a particularly significant weakness because it is so uncommon at the time of writing for team members on the ground to be adequately equipped to assess equity issues~\cite{posselt2020equity}.
Another significant challenge will be equipping teams to be genuinely willing to pause or wind down projects when that would lead to the best outcome because it can be a difficult and personally impactful decision to make. 
Furthermore, there is a risk of backlash and inertia derailing more equitable R\&D, regardless of the evidence.
Examples from the range of possible reasons for rejecting such signals might include attachment to one's work or the sunk cost fallacy.
Furthermore, processes we outline here are vulnerable to internal organizational discrimination against people based on identity, who are disproportionately likely to be forced out of organizations for calling attention to problems or injustices~\cite{nap2020promisingpractices,ahmed2021complaint}. 

Many of the individuals willing to embrace working towards more Equitable R\&D will require step-by-step guidance and resources relevant to their field before they can meaningfully get started and build broader organic engagement.
Even good-faith engagement and contributions can be hindered by human social network challenges. 
Many teams are likely to be hindered by a lack of connection to and understanding of communities that should be included. 
Some people will also have difficulty interacting with those communities in a manner that leads to desirable outcomes~\cite{porter2022intellectualhumility, scott1998seeinglikeastate,jackson2022disabilitydongle}.
In cases where individuals meaningfully commit to working towards more Equitable R\&D, many will remain ill equipped to build a diverse team or seek feedback, and 
even with an improved process some populations will still be excluded. 
There can even be conflicting access needs between groups as identity factors vary, so while our proposal has potential to lead to improvements, there is no complete `solution' for all situations.
We also recognize peoples' autonomy and the impossibility of representing all views. 
People’s lived experiences vary enormously, and population level discourse or trends do not represent all cases.
It is expected that even if the best possible project result and/or product is put out into the world, people are autonomous and have agency, so some might take it and make it their own in both positive and negative ways. 
Addressing every possible contingency is infeasible, so our framework aims to be cognizant and respectful of people's autonomy, and responsive to improve outcomes in ways that also mitigate negative outcomes where possible.
Additionally, while case studies are a valuable sample, the full set of world experiences are extremely diverse, cannot be fully captured, and shift over time~\cite{maza2017thinking}.
There are cases and views that we do not know we don't know, and thus we welcome constructive critique and seek iterative methodological refinement.
Another potential critique of this work is that aspects of it repackages and perhaps even co-opts well-understood and widely practiced methods that have been available for decades in various communities and fields~\cite{johnson2020undermining}.
Powerful actors might also dismiss the approach or co-opt the system themselves. 
The last two cases both have historical precedents in which harm is done through the process of engaging with good intentions~\cite{johnson2020undermining,maza2017thinking,wachter2017technicallywrong}.

\textbf{Workplace and Education Environment}
\label{subsec:workplace_education_environment}
A prerequisite to the development processes we describe here is a workplace environment that buys in and is willing to make coordinated focused effort~\cite{posselt2020equity} at multiple organizational levels (Fig. \ref{fig:scopes}).
It will not be effective in a toxic workplace environment, a detrimental problem that is dismayingly common and with proven persistence in both Industry and Academia~\cite{nap2020promisingpractices, nap2018sexualharassmentofwomen, posselt2020equity,johnson2020undermining,birhane2021algorithmic,spoon2023genderfacultyretention}.
For example, a National Academies Press Scientific Consensus report indicates ``substantial research demonstrates that implicit and explicit biases discourage women from entering STEMM careers or influence their decision to leave STEMM after beginning their careers. These factors include a spectrum of explicit and implicit biases, as well as structural and interpersonal interactions that impede women’s progress. These are factors across the career life cycle''~\cite{nap2020promisingpractices}.
The process we describe is also vulnerable to a powerful stakeholder, such as an Executive or Principal Investigator simply derailing~\cite{derailing2018} or squashing the system with their authority.
Shortcomings of and divergences between written and practiced policy is another common vulnerability~\cite{ahmed2021complaint} for all organizational processes.
New legislative frameworks and incentives might need to be developed to address such priorities, and we encourage legislators to advance and democratize deliberative processes in the sense of democratic governance, but specific legislative recommendations are out of scope for this work.

The actual internal development process and organization team should also operate as equitably as possible, with cognizance of each others humanity and proper coordination and accommodations for the needs of each team member~\cite{nap2020promisingpractices,dolmage2017academic,dignazio2020datafeminism}.
However, we recognize that individuals are human and meaningful changes can take time, so a pragmatic goal is to ensure ongoing iterative improvements in team capability for equity of interactions and the working environment.
Broader discourse on this topic is out of the scope of this work, so we refer the reader to recent scientific consensus reports on evidence based best practices with respect to Responsible Computing~\cite{nap2022responsiblecomputingresearch} Research, reducing the sexual harassment of women~\cite{nap2018sexualharassmentofwomen}, reducing the underrepresentation of women~\cite{nap2020promisingpractices} in STEMM, mentorship~\cite{nap2019mentorship}, and the Universal Design for Learning Guidelines~\cite{UDLguidelines22}. 
Each of these resources also considers identity as an important factor in their analysis and recommendations.

\textbf{Future Work}
Questions to consider for future work include:
How much engagement should there be across identities worldwide and what are the criteria to determine such decisions? 
How to shift more power to minoritized groups?
How to guide educating the team, perhaps through self study or a specific education package?
How to onboard people from minoritized groups and other groups?
How to start seeking community feedback in other countries or groups?
How to highly prioritize and integrate community feedback?
How to change budget priorities? 
How best to prioritize at different budgets/team sizes?
There is also potential for policy and legal changes at the level of institutions and government to more effectively mitigate potential concerns, such as required integration of independent ethics review steps and audits, or even a license to practice~\cite{pasquale2020newlawsofrobotics}.

\section{Conclusion}
We have proposed a framework that aims to serve as a step towards equitable agile research and development of `AI' and Robotics, so organizations can build the capability to examine their particular problem and choose the right tools to ensure more equitable outcomes to their project.
We draw connections between the topics of research, the methods being practiced, and people being included or excluded as part of the human process of developing research or products that contain `AI' or Robotics.
The applicability of particular tools in this toolkit will vary across projects and they should be chosen on a case by case basis.
We introduced concrete steps such as ethical and equity discussion components of meetings and retrospectives, participant-led project assessment scorecards, a participant-led governance model, the time blocks for general use, and the time blocks for building organizational capabilities through education and experimentation.
We hope this work will support projects and communities to thrive.
\newpage
\begin{acks}
We thank William Agnew, Anne Washington, Kari George, Di Wu, Christopher Dancy, Peter Schaldenbrand, industry and academic partners who provided feedback at the WeRobot 2023 conference, and those not listed here for the valuable feedback and insightful discussions during various stages of this work. 
We thank \citet{wilson2018agile} for their compelling visualization of the traditional Agile System Development lifecycle as well as their mobile device clinical Health (mHealth) Agile Development \& Evaluation Lifecycle. 
We adapted and expanded upon these under their \href{https://creativecommons.org/licenses/by/4.0/}{Creative Commons Attribution 4.0 License (CC BY 4.0)} to develop our Fig. \ref{fig:scrum_process}'s More Equitable Agile Research and Development Lifecycle. 
We thank the Council of Medical Specialty Societies (CMSS) and Patient-Led Research Collaborative (PLRC) for their brilliant Patient-led Research Scorecards~\cite{plrc2023scorecards,davisLongCOVIDMajor2023}, which we directly paraphrased and then expanded with permission from a medicine-specific assessment focus to scorecards capable of assessing a much broader range of application domains, identities, development artifacts, and licensing criteria. To the best of our knowledge, Sec. \ref{sec:scorecards} (the scorecards), and the latex code for their table formatting, are the only section modified with Large Language Model (LLMs) based tools, specifically ChatGPT and Bing Chat. 
Finally, the Researcher and Staff Support Scorecard in Sec. \ref{subsec:researcher_and_staff_support_scorecard} is our not just partially new, but completely new addition to the scorecards.
\end{acks}
\bibliographystyle{ACM-Reference-Format}
\bibliography{bibliography}

\newpage
\appendix

\section{Definitions}
\label{sec:definitions}
For inclusive working definitions of identity, race, ethnicity, sex, gender, data setting, and dissolution model in an `AI' and Robotics context see \citet{hundt2022robots_enact}.
Many definitions have multiple different perspectives and evolve over time, so our definitions are intended to serve as a useful starting point.

\begin{description}[leftmargin=0.0cm]
\item[Data Settings] ``Rather than talking about datasets, [data studies scholar Yanni Loukissas~\cite{loukissas2019all}] advocates that we talk about data settings—his term to describe both the technical and the human processes that affect what information is captured in the data collection process and how the data are then structured.''~\cite{dignazio2020datafeminism}

\item[Crip Time] %
``Recognizing how expectations of how long things should take account [of a range of] types of minds and bodies” so that we can `bend the clock' rather than bending bodies~[\cite{kafer2013feministqueercrip}]. Margaret Price suggests that crip time is the `flexible approach to normative time frames' ([\cite{price2011mad}]). Crip time has generally been interpreted as responsive: a way to impose critical delay through the refusal to follow strict schedules (schedules that might be normative, ableist, medically rehabilitative, and so on).'' -  \citet{dolmage2017academic} (p. 179)

\item[Universal Design for Learning (UDL)] ``An approach to curricula and teaching methods that strives to be more inclusive than American with Disabilities Act guidelines.''~\cite{nap2019mentorship} See also: CAST UDL Guidelines~\cite{UDLguidelines22}.

\item[Intellectual Humility] The ``meta-cognitive ability to recognize the limitations of one’s beliefs and knowledge [... which] might be particularly important for scientists for its role in enabling scientific progress''~\citet{porter2022intellectualhumility}. \citet{porter2021intellectualhumility} surveys intellectual humility definitions.

\item[Cultural Humility] ``An emerging idea that highlights the ability to maintain openness to others, engage in self-critique, redress power asymmetries, and work in partnerships to advocate for others.''~\cite{posselt2020equity}.

\item[Jidoka] ``The term Jidoka as used at Toyota means `to make the equipment or operation stop whenever an abnormal or defective condition arises'.''~\cite{SUGIMORIY1977toyotaKanban}
\end{description}

In this work ``identity'' collectively refers to factors such as race,
indigenous identity, physical ability, mental ability, age, national origin,
experience, cultural conventions, gender identity, LGBTQIA+ identity, etc.
Altogether, there is comparatively little investigation into ML and `AI' with
respect to identity factors when compared to algorithm performance on other metrics which can be problematic when misused
~\cite{dignazio2020datafeminism,buolamwini2018gender,birhane2021algorithmic,bender2021on}. 
For example, ``accuracy'' can leave segments of a large population completely excluded or subject to harm as `outliers'.

\section{Additional Details}
\label{sec:additional_details}
\subsection{Task Work Example of Equity Context and Human Override}
\label{subsec:task_work}
Task work such as Amazon Mechanical Turk is one of the myriad powerful examples of the way human overrides and an ethical goal of understanding and prioritizing marginalized perspectives is simultaneously fundamental to technical goals. 
\citet{gray2019ghost} and others~\cite{difallah2018demographics,hara2018a,difallah2018demographics,wu2023disabilityexpertise} have conducted research into task work. 
They have found that these positions can be very precarious, to the point where it can be difficult for task workers to take care of their bills, their family, and their mental health; journalists have described cases with a striking resemblance to colonialism~\cite{hao2022aicolonialismtaskwork}.
Task workers build varied expertise, and can even be engineers and PhDs~\cite{gray2019ghost}.

Task instructions can be very unclear and designed without functional feedback mechanisms, such that they are nearly impossible to complete. 
To handle this and other aspects of their job, many have created their own meeting webpages, facebook groups, forums, etc, independently of the company operating the platform. 
They discuss the outcomes of various tasks, how to submit or organize or label data in a way that will get them paid.
Side channel communication can variously improve the task results, undermine them, or even invalidate them without the knowledge of task hosts, depending on the case, purpose, requirements, and type of communication.
For example, if a task worker makes a guide that improves the data collection process, this will be beneficial to results. 
However, this labor will go uncredited without two way communication, despite their work being a contribution that might have been worthy of co-authorship had they done the same exact work in person as part of a research role. 
Invalid results are possible in across-user studies where aspects of the task are differentially hidden as part of an experiment, communication of critical details might break the experiment if task hosts neglect the way the human processes actually behave, and design accordingly. 

``Good for tech: Disability expertise and labor in China's artificial intelligence sector'' provides a window into labeling as source of empowerment and freedom; it elaborates on the complexities of labeler expertise that task hosts and AI researchers often do not understand, and even occasionally impede.
\citet{gray2019ghost} describes steps and approaches that mitigate these considerations for more effective data collection, including and supporting the task workers on teams, and creating interactive development processes.

\subsection{Method Adoption and Team Size}

Our method aims to scale from micro (1-9 people) up through large organizations (250+ people), as defined by the OECD~\cite{oecd2019sme}. The number and depth of processes can grow over time as the organization grows. 
Organizations with a small budget and team are expected to adopt our toolkit in a manner that is substantively different from those with a large budget allocation. 

People working on micro projects (1-9) can learn essentials of sociotechnical impacts~\cite{dignazio2020datafeminism}, tailor a project based on existing published resources, conduct quick assessments, adopt inclusive guidelines, and seek comments from different communities and affinitiy groups in person or online, provided legal and policy requirements are met such as Institutional Review Boards (IRBs).
Micro organizations tend to have the tightest resource limitations, but even small steps can make a big difference, for example the basics of Model Cards~\cite{mitchell2019model} take a few days to learn and can help develop an appropriate scope and plan for a project, and by carefully detailing the purpose of any data collected~\cite{gebru2018datasheets}, this can help prevent misuse. Artifacts (Sec. \ref{subsec:artifacts}) can be managed as simply as a plain text file in a code base or a versioned text document.
Detailed studies can be sourced from existing research, and new questions might be initially addressed by reaching out to experts or posting thoughtful questions to appropriate online affinity groups.

The resources in Sec. \ref{subsec:participatory_design} on Participatory design also detail a range of methods that are effective from micro (1-10 people) up through medium scale projects.
As projects increase in headcount through large sale projects (250+ people), additional systematization, adoption of more of the toolkit, more representative deliberative governance, and more comprehensive investigation of underlying issues becomes feasible. 
A detailed treatment is out of scope for this paper, so we recommend \citet{yacine2022datagovernance} for the governance of large scale `AI' projects.

One opportunity on new teams and projects is there is more flexibility to define an inclusive and participatory culture, and to create a diverse team. 
Existing Agile teams and projects have a body of knowledge about how to conduct adaptable project planning, so the new task and time elements can be easier to integrate on a scheduling basis.
This Agile process is designed for integration with many other processes such as datasheets~\cite{gebru2018datasheets}, model cards~\cite{mitchell2019model}, Audits~\cite{raji2020closing}, Diversity metrics~\cite{mitchell2020diversity}, EDAP ethical agile~\cite{kacianka2021designing}, and Participatory Design (Sec. \ref{subsec:participatory_design}).

\subsection{Financial Considerations}

\citet{viljoen2020democratic} argues that maximizing financial gain from or to data subjects misses the point, and that the purpose of data is to put people into person based relations with one another.
\citet{bessen2022ethicalaistartupcost} conducts a survey that outlines the interplay of economic pressures and `AI' ethics policies in startups that are under 10 years old. 
Caveated by the limitations of their methodology, they find: %
\begin{quote}
\leftskip=-0.3cm
\rightskip=-0.3cm
``that 58\% of these startups have established a set of `AI' principles. Startups with data-sharing relationships with high-technology firms or that have prior experience with privacy regulations are more likely to establish ethical `AI' principles and are more likely to take costly steps, like dropping training data or turning down business, to adhere to their ethical `AI' policies. Moreover, startups with ethical `AI' policies are more likely to invest in unconscious bias training, hire ethnic minorities and female programmers, seek expert advice, and search for more diverse training data. Potential costs associated with data-sharing relationships and the adherence to ethical policies may create tradeoffs between increased `AI' product competition and more ethical `AI' production.''
\end{quote}

\section{Participant-led Project Assessment Scorecards}
\label{sec:scorecards}

The scorecards are discussed in Sec. \ref{subsubsec:participant_led_scorecards} with the cards themselves below. Adapted from Patient-led Research Scorecards~\cite{plrc2023scorecards}.

\begin{spacing}{1.0}

\pagebreak
\subsection{Participant Burden Scorecard}

\begin{longtable}{p{0.75cm}p{\dimexpr\textwidth-0.75cm-3\tabcolsep-2\arrayrulewidth}}
\hline
\rowcolor{light-gray}
\multicolumn{2}{|c|}{\large\textbf{Participant Burden Scorecard}} \\
\hline
\endfirsthead

\multicolumn{2}{c}{{\bfseries \tablename\ \ref{tab:participant_burden_scorecard} -- continued from previous page}} \\
\hline
\rowcolor{light-gray}
\multicolumn{2}{|c|}{\large\textbf{Participant Burden Scorecard}} \\
\hline
\endhead

\hline
\multicolumn{2}{r}{{Continued on next page}} \\
\endfoot

\hline
\endlastfoot

\rowcolor{lighter-gray}
\multicolumn{1}{|c}{\textbf{Score}} & \multicolumn{1}{|c|}{\textbf{Accessible Engagement}} \\
\hline
-2 & \textbf{Non-collaboration:} Research \& Development organization dictates engagement avenues with no consideration of the participant population's access needs. Full participation may be impossible; carry a high time, effort, or monetary cost; or cause participants harm. \\
\hline
\rowcolor{lightest-gray}
-1 & \textbf{Minimal collaboration:} Research \& Development organization considers the participant population when designing engagement avenues, but rarely provides additional accommodations when requested. \\
\hline
0 & \textbf{Acceptable collaboration:} Research \& Development organization designs engagement avenues to offer sufficient time and accessibility for the participant population's needs, and provides individuals with additional accommodations upon request. \\
\hline
\rowcolor{lightest-gray}
1 & \textbf{Great collaboration:} Research \& Development organization designs engagement avenues to offer sufficient time and accessibility for the participant population's needs, ensures participants can easily request additional accommodations, and works with participants to co-design systemic updates in response to requests. \\
\hline
2 & \textbf{Ideal collaboration:} Participants co-create engagement avenues from the outset to ensure that full participation is accessible and minimally harmful across impacted sub-populations. \\
\hline

\rowcolor{lighter-gray}
\multicolumn{1}{|c}{\textbf{Score}} & \multicolumn{1}{|c|}{\textbf{Trauma Informed Practices}} \\
\hline
-2 & \textbf{Non-collaboration:} Research \& Development organization does not consider possible trauma burdens for participants. Participants may experience discrimination, hostility, new or recalled trauma, or other harms as a result of participation. No trauma-informed practices are in place, and participants receive no resources or support for the trauma caused by participation. \\
\hline
\rowcolor{lightest-gray}
-1 & \textbf{Minimal collaboration:} Research \& Development organization is aware of a possible trauma burden, but no systemic trauma-informed practices are in place, and participants receive no resources or support for their trauma. \\
\hline
0 & \textbf{Acceptable collaboration:} Research \& Development organization recognizes potential trauma burdens, and some trauma-informed practices are in place. Resources and support are provided to participants upon request. \\
\hline
\rowcolor{lightest-gray}
1 & \textbf{Great collaboration:} Research \& Development organization implements trauma-informed practices throughout the study, and collaborates with participants to co-design adjustments to those practices during the project. Requests for resources and support are honored at a systemic level for all participants. \\
\hline
2 & \textbf{Ideal collaboration:} A diverse array of participants, representative of the project's sub-populations, collaborates from the outset to co-create a safe, inclusive, mutually respectful environment; implement and adjust trauma-informed practices throughout the research process; and ensure all participants proactively receive sufficient, comprehensive resources and support. \\
\hline

\rowcolor{lighter-gray}
\multicolumn{1}{|c}{\textbf{Score}} & \multicolumn{1}{|c|}{\textbf{Responsiveness to Participants}} \\
\hline
-2 & \textbf{Non-collaboration:} No formal channels for participant input are established. Research \& Development organization does not address participant feedback, and may exclude or retaliate against participants who voice concerns. \\
\hline
\rowcolor{lightest-gray}
-1 & \textbf{Minimal collaboration:} Participants find channels for input to be unclear, difficult to access, or unsafe from retaliation. Participant feedback may be acknowledged, but rarely results in changes to the current project. \\
\hline
0 & \textbf{Acceptable collaboration:} Research \& Development organization creates clear, accessible, safe channels for participant input only after the research process has begun. Participant feedback is acknowledged, resulting in changes to analysis, presentation, or communication; and ad-hoc changes to the current project. \\
\hline
\rowcolor{lightest-gray}
1 & \textbf{Great collaboration:} Research \& Development organization creates clear, accessible, safe channels for participant input throughout the research process; acknowledges participant feedback; and establishes mechanisms for participants to co-design systemic changes to the current project. \\
\hline
2 & \textbf{Ideal collaboration:} Participants co-lead the project from end to end, including creating clear, accessible, safe channels for input, using that input to inform the research process, and acknowledging its impact. Members of the Research \& Development organization are excited about and fully engaged in participant collaboration. \\
\hline

\rowcolor{lighter-gray}
\multicolumn{1}{|c}{\textbf{Score}} & \multicolumn{1}{|c|}{\textbf{Compensation}} \\
\hline
-2 & \textbf{Non-collaboration:} Participants are compensated below market rate for their domain expertise and experience level, with no or limited options for when and how they are paid. Expenses, harm, and risk assumed from participation are not compensated. \\
\hline
\rowcolor{lightest-gray}
-1 & \textbf{Minimal collaboration:} Participants are compensated at market rate for their expertise and experience, with no or limited payment options. Expenses, harm, and risk are not compensated. \\
\hline
0 & \textbf{Acceptable collaboration:} Research \& Development organization sets participant compensation at market rate for their expertise and experience; and for anticipated expenses, harm, and risk. Multiple payment options are offered upfront. Requests for additional compensation and/or payment options are honored ad-hoc. \\
\hline
\rowcolor{lightest-gray}
1 & \textbf{Great collaboration:} Research \& Development organization sets participant compensation at or above market rate for their expertise and experience; and for anticipated expenses, harm, and risk. Multiple payment options are offered upfront. Requests for additional compensation and/or payment options result in systemic changes that benefit all participants. \\
\hline
2 & \textbf{Ideal collaboration:} Participants have decision-making roles in setting and adjusting compensation. Participants are compensated at or above market rate for their expertise and experience; and for anticipated expenses, harm, and risk; in the method and timing of their choice. Requests benefit all participants. Participants receive non-monetary compensation in the form of visibility, professional development, authorship, and awareness of their impact. \\
\hline

\end{longtable}
\label{tab:participant_burden_scorecard}

\pagebreak
\subsection{Participant/Partner Governance Scorecard}

\begin{longtable}{p{0.75cm}p{\dimexpr\textwidth-0.75cm-3\tabcolsep-2\arrayrulewidth}}
\hline
\rowcolor{light-gray}
\multicolumn{2}{|c|}{\large\textbf{Participant/Partner Governance Scorecard}} \\
\hline
\endfirsthead

\multicolumn{2}{c}{{\bfseries Table continued from previous page}} \\
\hline
\rowcolor{light-gray}
\multicolumn{2}{|c|}{\large\textbf{Participant/Partner Governance Scorecard}} \\
\hline
\endhead

\hline
\multicolumn{2}{r}{{Continued on next page}} \\
\endfoot

\hline
\endlastfoot

\rowcolor{lighter-gray}
\multicolumn{1}{|c}{\textbf{Score}} & \multicolumn{1}{|c|}{\textbf{Meaningful Decision making between groups}} \\
\hline
-2 & \textbf{Non-collaboration:} Decision-making for significant decisions (funding, project design, publication, etc.) is not communicated transparently and/or the Research \& Development organization decides the decision-making process without participant input. \\
\hline
\rowcolor{lightest-gray}
-1 & \textbf{Minimal collaboration:} Decision-making process for significant decisions (funding, project design, publication, etc.) is not communicated and/or agreed upon. Participants have limited or not meaningful decision-making power. \\
\hline
0 & \textbf{Acceptable collaboration:} Decision-making process for significant decisions (funding, project design, publication, etc.) is well communicated and agreed upon between participants and the Research \& Development organization. \\
\hline
\rowcolor{lightest-gray}
1 & \textbf{Great collaboration:} Decision-making for significant decisions (funding, project design, publication, etc.) is well communicated and agreed upon between the participant and partner group, with deference given to the participant group. \\
\hline
2 & \textbf{Ideal collaboration:} Decision-making for significant decisions (funding, project design, publication, etc.) is well communicated and agreed upon between the participant and partner group, with deference given to the participant group with sufficient support to make the decisions. \\
\hline

\rowcolor{lighter-gray}
\multicolumn{1}{|c}{\textbf{Score}} & \multicolumn{1}{|c|}{\textbf{Accountability between groups}} \\
\hline
-2 & \textbf{Non-collaboration:} There is a lack of understanding of the rules of engagement/culture between groups with no written agreement and no defined consequences for not following through. \\
\hline
\rowcolor{lightest-gray}
-1 & \textbf{Minimal collaboration:} There is an understanding of the rules of engagement/culture but no written agreement and/or defined consequences for not following through between groups. \\
\hline
0 & \textbf{Acceptable collaboration:} There is a shared understanding and written agreement of the rules of engagement/culture with defined consequences for not following through between groups. \\
\hline
\rowcolor{lightest-gray}
1 & \textbf{Great collaboration:} Shared understanding and written agreement of the rules of engagement/culture with defined consequences for not following through between groups. Deference is given to participant groups to define the engagement. \\
\hline
2 & \textbf{Ideal collaboration:} Shared understanding and written agreement of the rules of engagement/culture with defined consequences for not following through between groups. Deference is given to participant groups to define the engagement with sufficient support. \\
\hline

\end{longtable}
\label{tab:participant_governance_scorecard}

\pagebreak
\subsection{Research \& Development Organization Readiness Scorecard}

\begin{longtable}{p{0.75cm}p{\dimexpr\textwidth-0.75cm-3\tabcolsep-2\arrayrulewidth}}
\hline
\rowcolor{light-gray}
\multicolumn{2}{|c|}{\large\textbf{Research \& Development Organization Readiness Scorecard}} \\
\hline
\endfirsthead

\multicolumn{2}{c}{{\bfseries \tablename\ \thetable{} -- continued from previous page}} \\
\hline
\rowcolor{light-gray}
\multicolumn{2}{|c|}{\large\textbf{Research \& Development Organization Readiness Scorecard}} \\
\hline
\endhead

\hline
\multicolumn{2}{r}{{Continued on next page}} \\
\endfoot

\hline
\endlastfoot

\rowcolor{lighter-gray}
\multicolumn{1}{|c}{\textbf{Score}} & \multicolumn{1}{|c|}{\textbf{Recognition of Biases}} \\
\hline
-2 & \textbf{Non-collaboration:} Research \& Development organization does not recognize bias and ignores feedback from participants. \\
\hline
\rowcolor{lightest-gray}
-1 & \textbf{Minimal collaboration:} Research \& Development organization has limited awareness of its own biases and listens to some feedback from participants. \\
\hline
0 & \textbf{Acceptable collaboration:} Research \& Development organization is aware of its own biases, is open to feedback from participants, and implements some of the feedback. \\
\hline
\rowcolor{lightest-gray}
1 & \textbf{Great collaboration:} Research \& Development organization is aware of its own biases, is open to feedback from participant groups, and actively iterates on feedback given. \\
\hline
2 & \textbf{Ideal collaboration:} Research \& Development organization is aware of its own biases and is open to listening to feedback from participant groups. It actively iterates on feedback given. Other participant groups can attest to a positive working relationship. The Research \& Development organization has a systemic process for accepting input from participants and participant groups. \\
\hline

\rowcolor{lighter-gray}
\multicolumn{1}{|c}{\textbf{Score}} & \multicolumn{1}{|c|}{\textbf{Collaboration Process}} \\
\hline
-2 & \textbf{Non-collaboration:} Research \& Development organization has no dedicated infrastructure for collaborating with participants. \\
\hline
\rowcolor{lightest-gray}
-1 & \textbf{Minimal collaboration:} Research \& Development organization has minimal resources/infrastructure for collaborating with participants. \\
\hline
0 & \textbf{Acceptable collaboration:} Research \& Development organization has dedicated some resources and infrastructure for collaborating with participants (e.g., participant panels); has at least one coordinating personnel focused on meeting the participant group's needs; conducts limited training to build skills to engage with participants. \\
\hline
\rowcolor{lightest-gray}
1 & \textbf{Great collaboration:} Research \& Development organization has an established infrastructure and process for collaborating and co-designing with participants, including at least one dedicated person focused on meeting the participant group's needs and advocating to the rest of the Research \& Development organization. It conducts routine training to build skills to engage with participants. \\
\hline
2 & \textbf{Ideal collaboration:} Research \& Development organization has an established infrastructure and process for collaborating with participants that has been vetted by other participants/participant groups. It has at least one dedicated person who is focused on meeting the participant group's needs. The partner is recognized as a participant ally vetted by other participants and participant groups with a background in justice as it applies to impacted identities (e.g. disability justice, racial justice, etc.). The organization conducts extensive training on meaningful engagement with participants. \\
\hline

\rowcolor{lighter-gray}
\multicolumn{1}{|c}{\textbf{Score}} & \multicolumn{1}{|c|}{\textbf{Knowledge of Impacted Identities}} \\
\hline
-- & \textit{Please provide a score for each impacted identity about which you wish to provide feedback:
Race, physical disability, cognitive disability, age, national origin, cultural, gender, LGBTQIA+ status, wealth status, income status, an intersection of multiple identities, or specify one or more of your own identities.}\\
\hline
-2 & \textbf{Non-collaboration:} Research \& Development organization has no knowledge/experience with the impacted identities. \\
\hline
\rowcolor{lightest-gray}
-1 & \textbf{Minimal collaboration:} Research \& Development organization has minimal knowledge/experience (less than one year) with the impacted identities. \\
\hline
0 & \textbf{Acceptable collaboration:} Research \& Development organization has at least one year worth of knowledge/experience with the impacted identities. \\
\hline
\rowcolor{lightest-gray}
1 & \textbf{Great collaboration:} Research \& Development organization has more than one year worth of knowledge/experience of the impacted identities. \\
\hline
2 & \textbf{Ideal collaboration:} Research \& Development organization has extensive knowledge and direct experience with the impacted identities, and those with knowledge are in decision-making roles. The Research \& Development organization has a systemic way to keep on top of information from the participant community as well as the latest research findings. \\
\hline

\end{longtable}
\label{tab:r_and_d_readiness_scorecard}

\pagebreak

\subsection{Integration Into Research Process Scorecard}
\label{subsec:integration_into_research_process_scorecard}

\begin{longtable}{p{0.75cm}p{\dimexpr\textwidth-0.75cm-3\tabcolsep-2\arrayrulewidth}}
\hline
\rowcolor{light-gray}
\multicolumn{2}{|c|}{\large\textbf{Integration Into Research Process Scorecard}} \\
\hline
\endfirsthead

\multicolumn{2}{c}{{\bfseries \tablename\ \thetable{} -- continued from previous page}} \\
\hline
\rowcolor{light-gray}
\multicolumn{2}{|c|}{\large\textbf{Integration Into Research Process Scorecard}} \\
\hline
\endhead

\hline
\multicolumn{2}{r}{{Continued on next page}} \\
\endfoot

\hline
\endlastfoot

\rowcolor{lighter-gray}
\multicolumn{1}{|c}{\textbf{Score}} & \multicolumn{1}{|c|}{\textbf{Hypothesis Generation}} \\
\hline
-2 & \textbf{Non-collaboration:} Research goals are siloed from participants' priorities. Participants' questions and experiences are not included and/or are dismissed when generating research hypotheses. \\
\hline
\rowcolor{lightest-gray}
-1 & \textbf{Minimal collaboration:} Research goals attempt to involve participants' priorities but are limited by communication or collaboration. Participants' inquiries and lived experiences are rarely included when generating research hypotheses. Participants may have suggested the research question with no further involvement. \\
\hline
0 & \textbf{Acceptable collaboration:} Research goals take into account participants' priorities. Participants' inquiries and lived experiences are included when generating research hypotheses. \\
\hline
\rowcolor{lightest-gray}
1 & \textbf{Great collaboration:} Research goals proactively address participants' priorities with sufficient ongoing collaboration. Participant organization's inquiries and lived experiences are included when generating research hypotheses. Participant organizations work with participants to co-design research hypotheses. \\
\hline
2 & \textbf{Ideal collaboration:} Research goals are based on participants' priorities and co-written by participant organization or participant-researchers. Participants' inquiries and lived experiences share equal weight with Research \& Development organization's interests when generating research hypotheses. \\
\hline

\rowcolor{lighter-gray}
\multicolumn{1}{|c}{\textbf{Score}} & \multicolumn{1}{|c|}{\textbf{Project Design}} \\
\hline
-2 & \textbf{Non-collaboration:} Research \& Development organization does not include participants in the project design process. Participants do not have the opportunity to provide input on the project design. Participant groups are utilized for recruitment purposes only, if at all. \\
\hline
\rowcolor{lightest-gray}
-1 & \textbf{Minimal collaboration:} Research \& Development organization does not include participants in the project design process. Participants may be invited to review the project design, but feedback is rarely incorporated, and no functioning accountability system is in place. \\
\hline
0 & \textbf{Acceptable collaboration:} Select participant voices are approached to inform the project design. Participants are invited to review the project design and have an impact on the project design. \\
\hline
\rowcolor{lightest-gray}
1 & \textbf{Great collaboration:} Participant organizations and their community's input are proactively invited to help inform the project design. Participant organizations are invited to co-design and review the project design, and participant feedback changes the project design. \\
\hline
2 & \textbf{Ideal collaboration:} Project design is co-written and reviewed by a diverse array of participant-researchers representative of the project's sub-populations. If applicable, protocol testing is done by the participant community. \\
\hline

\rowcolor{lighter-gray}
\multicolumn{1}{|c}{\textbf{Score}} & \multicolumn{1}{|c|}{\textbf{Analysis}} \\
\hline
-2 & \textbf{Non-collaboration:} Participants do not have input on what data to prioritize for analysis and methods of analysis. \\
\hline
\rowcolor{lightest-gray}
-1 & \textbf{Minimal collaboration:} Participants are asked to review manuscript drafts and prototypes, as applicable, but have little say in what data to prioritize for analysis and methods of analysis. \\
\hline
0 & \textbf{Acceptable collaboration:} Participants are involved in interpreting data and carrying out analysis in some capacity. \\
\hline
\rowcolor{lightest-gray}
1 & \textbf{Great collaboration:} Participants or participant organizations are invited and involved in interpreting data and carrying out analysis anywhere in the project. \\
\hline
2 & \textbf{Ideal collaboration:} Participant-researchers co-lead on the interpretation and analysis and/or work concurrently with the partner organization's research team to carry out the analysis. \\
\hline
\rowcolor{lighter-gray}
\multicolumn{1}{|c}{\textbf{Score}} & \multicolumn{1}{|c|}{\textbf{Products, Artifacts, and/or Publications}} \\
\hline
-2 & \textbf{Non-collaboration:} Project results are inaccessible to participants and/or behind an academic paywall. Findings are not communicated in lay terms and/or products are not available. Data with a probable risk of an overall negative impact is released anyway. \\
\hline
\rowcolor{lightest-gray}
-1 & \textbf{Minimal collaboration:} Research \& Development organization summarizes findings in lay terms, but project results are inaccessible to participants, are behind avoidable cost barriers such as an academic paywall. Data with a foreseeable risk of an overall negative impact is released anyway, with consent but no discussion of the risks. \\
\hline
0 & \textbf{Acceptable collaboration:} Project results are freely accessible to participants and the public. Findings are summarized in lay terms in ways that are informative to the participant and impacted populations. Physical and Digital artifacts as well as licenses are available on fair, reasonable, and non-discriminatory (FRAND) terms. Data with a potential net negative impact is released with minimally informed consent. \\
\hline
\rowcolor{lightest-gray}
1 & \textbf{Great collaboration:} Project results are freely accessible to participants and the public. Findings are summarized in lay terms and are actively disseminated to the participant and impacted populations. Participant-researchers co-write the interpretation and analysis. Good faith dissenting assessments are welcome. Physical and Digital artifacts as well as licenses are available on fair, reasonable, and non-discriminatory (FRAND) terms. Potential net negative impacts from data release considered before data release with informed consent. \\
\hline
2 & \textbf{Ideal collaboration:} Project results are freely accessible to participants and the public. Findings are summarized in lay terms and are actively disseminated to the participant and impacted populations. Participant organizations invite participants to co-write findings and reports. A channel of communication is available for participants to ask questions of the Research \& Development organization. Good faith dissenting assessments are welcome. Physical and Digital artifacts are available for free or at cost with open licenses (where applicable) in a manner better than fair, reasonable, and non-discriminatory (FRAND) terms. Potential net negative impacts from data release are carefully mitigated, and data is not released if the risk is too high, or substantive steps are taken to inform participants and obtain full informed consent. \\
\hline

\rowcolor{lighter-gray}
\multicolumn{1}{|c}{\textbf{Score}} & \multicolumn{1}{|c|}{\textbf{Attribution}} \\
\hline
-2 & \textbf{Non-collaboration:} Participants' work is attributed to others and/or participants are not attributed at all. \\
\hline
\rowcolor{lightest-gray}
-1 & \textbf{Minimal collaboration:} Participants are listed as being involved without a description of how they were involved. Participants were not consulted on how they prefer to be attributed. \\
\hline
0 & \textbf{Acceptable collaboration:} Participants are acknowledged/credited in major public-facing communication (press, announcements, papers), to the extent that participants wish to be named. Participants were consulted on how they prefer to be attributed. \\
\hline
\rowcolor{lightest-gray}
1 & \textbf{Great collaboration:} Participant group is credited in all public-facing communication and included as authors on papers or products, to the extent that the participant group wishes to be named. Participant group was consulted on how they prefer to be attributed. \\
\hline
2 & \textbf{Ideal collaboration:} Participants are acknowledged specifically for what they did throughout the engagement process, are credited in all public-facing communication, and included as authors on papers or products, to the extent that the participant group wishes to be named. Participant group was consulted on how they prefer to be attributed. \\
\hline

\end{longtable}
\label{tab:integration_into_research_process_scorecard}

\pagebreak

\subsection{Researcher and Staff Support Scorecard}
\label{subsec:researcher_and_staff_support_scorecard}

\begin{longtable}{p{0.75cm}p{\dimexpr\textwidth-0.75cm-3\tabcolsep-2\arrayrulewidth}}
\hline
\rowcolor{light-gray}
\multicolumn{2}{|c|}{\large\textbf{Researcher and Staff Support Scorecard}} \\
\hline
\endfirsthead

\multicolumn{2}{c}{{\bfseries Table continued from previous page}} \\
\hline
\rowcolor{light-gray}
\multicolumn{2}{|c|}{\large\textbf{Researcher and Staff Support Scorecard}} \\
\hline
\endhead

\hline
\multicolumn{2}{r}{{Continued on next page}} \\
\endfoot

\hline
\endlastfoot

\rowcolor{lighter-gray}
\multicolumn{1}{|c}{\textbf{Score}} & \multicolumn{1}{|c|}{\textbf{Accessibility}} \\
\hline
-2 & \textbf{Non-collaboration:} Support services and resources are difficult to find, access, or use. Researchers and staff face barriers such as lack of information, complex procedures, technical issues, or limited availability. \\
\hline
\rowcolor{lightest-gray}
-1 & \textbf{Minimal collaboration:} Support services and resources are somewhat accessible but require significant effort or time from researchers and staff. Researchers and staff encounter some challenges such as insufficient information, unclear procedures, occasional glitches, or limited options. \\
\hline
0 & \textbf{Acceptable collaboration:} Support services and resources are adequately accessible but could be improved. Researchers and staff can access them with reasonable effort and time. Researchers and staff experience few difficulties such as outdated information, inconsistent procedures, minor errors, or limited flexibility. \\
\hline
\rowcolor{lightest-gray}
1 & \textbf{Great collaboration:} Support services and resources are easily accessible and user-friendly. Researchers and staff can access them with minimal effort and time. Researchers and staff have no major problems such as inaccurate information, confusing procedures, frequent failures, or rigid policies. \\
\hline
2 & \textbf{Ideal collaboration:} Support services and resources are highly accessible and customized. Researchers and staff can access them with ease and convenience. Researchers and staff have a positive experience such as comprehensive information, streamlined procedures, reliable performance, or adaptive solutions. \\
\hline

\rowcolor{lighter-gray}
\multicolumn{1}{|c}{\textbf{Score}} & \multicolumn{1}{|c|}{\textbf{Work Life Balance}} \\
\hline
-2 & \textbf{Non-collaboration:} Researchers and staff have poor work-life balance. They face excessive workload, stress, or pressure that negatively affect their health, well-being, or personal life. They have no flexibility or autonomy in their work schedule or location, and cannot work remotely if needed. They struggle to complete their work efficiently and effectively, are dissatisfied with their work quality and outcomes, and receive no or negative feedback or recognition for their work. \\
\hline
\rowcolor{lightest-gray}
-1 & \textbf{Minimal collaboration:} Researchers and staff have low work-life balance. They face high workload, stress, or pressure that moderately affect their health, well-being, or personal life. They have limited flexibility or autonomy in their work schedule or location, and face barriers to work remotely if needed. They find it hard to complete their work efficiently and effectively, are unhappy with their work quality and outcomes, and receive little or mixed feedback or recognition for their work. \\
\hline
0 & \textbf{Acceptable collaboration:} Researchers and staff have adequate work-life balance. They face manageable workload, stress, or pressure that slightly affect their health, well-being, or personal life. They have some flexibility or autonomy in their work schedule or location, and can work remotely if needed with some support. They are able to complete their work efficiently and effectively, are satisfied with their work quality and outcomes, and receive adequate or positive feedback or recognition for their work. \\
\hline
\rowcolor{lightest-gray}
1 & \textbf{Great collaboration:} Researchers and staff have a strong work-life balance. They face balanced workload, stress, or pressure that enhance their health, well-being, or personal life. They have a lot of flexibility or autonomy in their work schedule or location, and can work remotely if needed with ease. They can complete their work efficiently and effectively with ease, are proud of their work quality and outcomes, and receive frequent or constructive feedback or recognition for their work. \\
\hline
2 & \textbf{Ideal collaboration:} Researchers and staff have optimal work-life balance. They face optimal workload, stress, or pressure that boost their health, well-being, or personal life. They have full flexibility or autonomy in their work schedule or location, and can work remotely if needed with creativity. They excel at completing their work efficiently and effectively with creativity, are delighted with their work quality and outcomes, and receive consistent or exceptional feedback or recognition for their work. \\
\hline

\rowcolor{lighter-gray}
\multicolumn{1}{|c}{\textbf{Score}} & \multicolumn{1}{|c|}{\textbf{Attribution}} \\
\hline
-2 & \textbf{Non-collaboration:} Researchers' and staff's work is attributed to others and/or researchers and staff are not attributed at all. Researchers and staff face discrimination, exclusion, or exploitation based on their identity, role, or contribution. Researchers, staff scientists, staff engineers, service staff, contractors, family, students, assistants, and colleagues who contributed with material, editing, logistical, research, or in other capacities are not consulted on how they prefer to be attributed, recognized, or credited. There may be uncredited ghostwriters or other uncredited people who made research contributions. \\
\hline
\rowcolor{lightest-gray}
-1 & \textbf{Minimal collaboration:} Researchers' and staff's work is acknowledged in a generic or vague way, without specifying their individual or collective contributions. Researchers and staff are not given equal opportunities or recognition based on their identity, role, or contribution. Researchers, staff scientists, staff engineers, service staff, contractors, family, students, assistants, and colleagues who contributed with material, editing, logistical, research, or in other capacities are rarely consulted on how they prefer to be attributed, recognized, or credited. \\
\hline
0 & \textbf{Acceptable collaboration:} Researchers' and staff's work is acknowledged in a fair and respectful way, with some indication of their individual or collective contributions. Researchers and staff are given equal opportunities and recognition based on their identity, role, or contribution. Researchers, staff scientists, staff engineers, service staff, contractors, family, students, assistants, and colleagues who contributed with material, editing, logistical, research, or in other capacities are consulted on how they prefer to be attributed, recognized, or credited. \\
\hline
\rowcolor{lightest-gray}
1 & \textbf{Great collaboration:} Researchers' and staff's work is acknowledged in a detailed and specific way, with clear indication of their individual or collective contributions. Researchers and staff are given equal opportunities and recognition based on their identity, role, or contribution. Researchers and staff are involved in the decision-making process regarding the attribution of credit. Researchers, staff scientists, staff engineers, service staff, contractors, family, students, assistants, and colleagues who contributed with material, editing, logistical, research, or in other capacities are consulted on how they prefer to be attributed, recognized, or credited. \\
\hline
2 & \textbf{Ideal collaboration:} Researchers' and staff's work is acknowledged in a comprehensive and inclusive way, with explicit indication of their individual or collective contributions. Researchers and staff are given equal opportunities and recognition based on their identity, role, or contribution. Researchers and staff share the decision-making power regarding the attribution of credit. Researchers, staff scientists, staff engineers, service staff, contractors, family, students, assistants, bystanders, and practitioners who provide their expertise in the field (e.g., nurses, medical care staff, practitioners in any application industry), task workers (e.g., Amazon Mechanical Turk), and colleagues who contributed with ideas, material, editing, logistical, research, or in other capacities are consulted on how they prefer to be attributed, recognized, or credited. The attribution of credit is transparent, accountable, and respectful of the diversity and dignity of all contributors. Everyone involved is credited specifically for what they did throughout the engagement process (e.g., providing encouragement, funding, transportation, data analysis, etc.), credited in all public-facing communication (paper, speech, website, social media), and included as authors or acknowledgments on papers or products when appropriate, with a statement of their significance. \\
\hline

\end{longtable}
\label{tab:researcher_and_staff_support_scorecard}

\subsection{Additional Details}

\begin{longtable}{p{0.75cm}p{\dimexpr\textwidth-0.75cm-3\tabcolsep-2\arrayrulewidth}}
\hline
\rowcolor{light-gray}
\multicolumn{2}{|c|}{\large\textbf{Additional Details (optional)}} \\
\hline
\endfirsthead
\rowcolor{light-gray}
\multicolumn{2}{|c|}{\large\textbf{Please optionally provide additional notes or information we should be aware of below.}} \\
\hline
\endhead

\hline
\multicolumn{2}{r}{{Continued on next page}} \\
\endfoot

\hline
\endlastfoot

\rowcolor{lighter-gray}
\multicolumn{1}{|c|}{\textbf{The space below is provided for you to write clarifications, notes, thoughts, or other information we should know.}} \\
\hline
 \\\\\\\\\\\\\\\\
\hline

\hline

\end{longtable}
\label{tab:participant_governance_scorecard}
\end{spacing}

\end{document}